\documentclass{article}

\usepackage{arxiv}

\usepackage{url}            
\usepackage{booktabs}       
\usepackage{amsfonts}       
\usepackage{nicefrac}       
\usepackage{microtype}      
\usepackage{xcolor}         
\usepackage{bm}
\usepackage{array}
\usepackage[utf8]{inputenc} 
\usepackage[T1]{fontenc}    
\usepackage{epsfig}
\usepackage{epstopdf}
\usepackage{graphics}
\usepackage{graphicx}
\usepackage[normalem]{ulem}
\usepackage{float}
\usepackage{colortbl}
\usepackage{subfigure}
\usepackage{multirow}
\usepackage{diagbox}
\usepackage{bbding}
\usepackage{mathrsfs}
\usepackage{amsmath} 
\usepackage{amsthm}
\usepackage{longtable}
\usepackage{makecell}
\usepackage{enumitem}
\usepackage{color}
\usepackage{algorithm}
\usepackage{algorithmic}
\usepackage{caption}

\newcolumntype{L}[1]{>{\raggedright\let\newline\\\arraybackslash\hspace{0pt}}m{#1}}
\newcolumntype{C}[1]{>{\centering\let\newline  \\\arraybackslash\hspace{0pt}}m{#1}}
\newcolumntype{R}[1]{>{\raggedleft\let\newline \\\arraybackslash\hspace{0pt}}m{#1}}

\title{Urban Generative Intelligence (UGI): A Foundational Platform for Agents in Embodied City Environment}
\author{Fengli Xu\thanks{The first three authors contribute to this work equally.} \quad Jun Zhang\footnotemark[1]  \quad  Chen Gao\footnotemark[1] \quad Jie Feng \quad Yong Li \\
Tsinghua University, Beijing, China \\
\texttt{\{fenglixu, chgao96, liyong07\}@tsinghua.edu.cn}
}

\renewcommand{\shorttitle}{Urban Generative Intelligence}

\begin{document}
\maketitle

\begin{abstract}

Urban environments, characterized by their complex, multi-layered networks encompassing physical, social, economic, and environmental dimensions, face significant challenges in the face of rapid urbanization. These challenges, ranging from traffic congestion and pollution to social inequality, call for advanced technological interventions. Recent developments in big data, artificial intelligence, urban computing, and digital twins have laid the groundwork for sophisticated city modeling and simulation. However, a gap persists between these technological capabilities and their practical implementation in addressing urban challenges in an systemic-intelligent way. This paper proposes Urban Generative Intelligence (UGI), a novel foundational platform integrating Large Language Models (LLMs) into urban systems to foster a new paradigm of urban intelligence. UGI leverages CityGPT, a foundation model trained on city-specific multi-source data, to create embodied agents for various urban tasks. These agents, operating within a textual urban environment emulated by city simulator and urban knowledge graph, interact through a natural language interface, offering an open platform for diverse intelligent and embodied agent development. This platform not only addresses specific urban issues but also simulates complex urban systems, providing a multidisciplinary approach to understand and manage urban complexity. This work signifies a transformative step in city science and urban intelligence, harnessing the power of LLMs to unravel and address the intricate dynamics of urban systems.
The code repository with demonstrations will soon be released here \url{https://github.com/tsinghua-fib-lab/UGI}.

\end{abstract}

\maketitle

\section{Introduction} 

Urban are complex systems with dynamic and multi-layered networks encompassing physical elements (buildings, roads, infrastructure), social structures (population distribution, organizations, culture), economic activities (industry, services, commerce), and environmental factors (natural resources, ecosystems, climate change)~\cite{batty2013new, berkowitz2003understanding, bettencourt2021introduction}. This intricate interplay creates uncertainty and dynamism, reflecting the complex interactions between human activities and the urban environment~\cite{bettencourt2021introduction}. Each individual, community, and organization within these systems is interconnected, influencing the city's overall characters and functionality~\cite{lyon2011community}. The primary challenge in urban complex systems is balancing economic growth, social welfare, and environmental sustainability amid rapid urbanization, and it faces critical challenges including traffic congestion, environmental pollution, resource scarcity, and infrastructure strain, all exacerbated by rapid urbanization~\cite{zheng2014urban}. Besides, social inequality and housing issues further impact residents' quality of life. The growing threats of climate change, such as extreme weather and rising sea levels, add to these challenges, highlighting the urgent need for solutions. Addressing these prominent urban issues is essential to ensure sustainable, equitable urban development and maintain the vitality of cities in a rapidly evolving global context~\cite{acuto2018building}.

In order to address above problems, the recent technological evolution began with big data, which provided rich and variety urban information~\cite{zhang2017understanding}. The complexity of this data necessitated the development of artificial intelligence (AI) for effective description, prediction and management~\cite{zheng2023spatial}. This synergy led to urban computing\cite{zheng2014urban}, which applies AI to big data for urban problem-solving. Building on this, digital twins~\cite{batty2018digital} and city simulations~\cite{zhang2022mirage,zhang2023city} emerged, creating virtual models of cities using real-time data and AI analysis. These technologies represent a progression from data collection (big data), to data analysis (AI), to application (urban computing), and finally to advanced simulation and modeling (digital twins and city simulations). Each stage builds upon the last, offering increasingly sophisticated tools for tackling urban complexities. However, despite these advancements, the ability of these technologies to systematically address the complexity of urban issues lies in the gap between technological capabilities and practical implementation. This shortfall implies that while these technologies are invaluable tools, they cannot yet comprehend and address the intricate and systemic challenges cities face, necessitating further advancements in AI and computational models to achieve more intelligent system-wide urban solutions.

The advent of artificial general intelligence, particularly large language models (LLMs), as a form of human-like intelligence presents a transformative opportunity in addressing urban challenges~\cite{fjelland2020general, wei2022emergent, brown2020language}. LLMs demonstrate emergent intelligence capabilities, mimicking human cognitive process to analyze and reason vast datasets. The human-like intelligence of LLMs can significantly contribute to the intelligence of urban systems by providing deep and context-aware insights. They can identify patterns, sentiments, and trends within urban discourse, offering a more nuanced understanding of the complex social and economic dynamics at play in urban environments, which further supports smart decision-making in urban planning and policy formulation. Thus, the integration of LLMs into urban systems holds great promise for finding comprehensive solutions to the multifaceted challenges in cities, which will push the urban technology to the next stage of urban intelligence.

In this paper, we propose Urban Generative Intelligence (UGI), a foundational platform that fosters the emergence, evolution, and application of generative intelligence in urban space and fuels the transition to smart future city. 
The UGI platform is built on top of an open digital infrastructure that consists of the UrbanKG knowledge graph~\cite{liu2023urbankg} and a city simulator engine~\cite{zhang2022mirage}. 
This infrastructure is capable of simulating realistic urban interactions based on multi-source urban data and providing embodied feedback to intelligent agents, setting it apart from existing sandboxes~\cite{park2023generative} or virtual environments~\cite{duncan2011minecraft}.
We design a standard language interface to expose the access to this digital infrastructure, facilitating the easy plugin of language models and the development of generative and embodied agents.
To create a foundation model for urban problems, we pre-train a LLM called CityGPT on general text corpus and high-quality city-specific data, including general text corpus, including domain-specific urban knowledge, task-solving process data and so on. Using CityGPT as a generative intelligence core, we propose a general framework of creating embodied agents for various urban tasks, such as planning transportation system, assisting policy-making, and simulating urban life and socioeconomic interactions~\cite{gao2023large}. 
Moreover, this general framework can be easily adapted to build customized agents, fostering the emergence of diverse intelligent agents supporting various aspects of urban intelligence. 
Through the realistic embodied feedback provided by digital infrastructure and the interactions with other agents, the LLM-empowered agents are able to learn from the environment, develop their own understanding, and further evolve their intelligence to deal with complicate urban tasks. Leveraging the CityGPT and various embodied agents, the UGI platform can help solve urban issues, support a wide range of urban applications, and explore new urban forms. These components collectively form the Foundation Platform of UGI, facilitating the emergence and advancement of generative intelligence in urban space.

The contributions of this paper can be summarized as the following aspects.

\begin{itemize}

\item We are the first to propose an open digital infrastructure for embodied urban environment simulation, which provides realistic feedback of urban experience via a natural language interface. It leverages UrbanKG knowledge graph and city simulator to provide textual feedback that can enable generative intelligence in urban space. 

\item We design and implement a foundation model for city problems, called \emph{CityGPT}. It is continuously pre-trained from general foundation model to incorporate urban knowledge extracted from text corpus, and then supervised fine-tuned to induce urban intelligence ability with domain-specific data.

\item We propose a general framework for generative city agents, which releases the generative intelligence of CityGPT in various urban tasks. We also propose several successful design cases for: a) simulation agents of physical mobility, economy activity, and social interaction in city; b) decision making agents that can serve as personal assistance of location recommendation and schedule planning.  

\item We introduce an evaluation framework to validate the performance of foundation model and generative city agents, quantifying urban generative intelligence as the levels of mastering knowledge, simple reasoning, and planing and decision making. It provides a standard and reference for the development of urban intelligence.
 
\end{itemize}

\section{Related Work}

\subsection{Complex Urban System}
Cities have long been viewed as a complex system of interconnected humans, things and space~\cite{batty2013new}.
Extensive research attentions are devoted to review the universal patterns of various statistics in complex urban system, such as city size and morphology shape~\cite{batty2008size}. Specifically, previous works find cities exhibit typical fractal morphology~\cite{batty1994fractal}, contradicting the common practice in urban planing, but is a symbolic feature of complex system. Besides, a large body of literature is devoted to reveal the scaling laws in urban space~\cite{bettencourt2007growth}. They find universal patterns of super-linear growth of economy, innovation and crimes, and sub-linear growth of infrastructure investment as city size increases. Previous works build theoretical framework to explain these scaling laws from the interaction mechanisms of urban agents~\cite{bettencourt2013origins}, which root in the increased interaction frequency in the compact urban space of large cities~\cite{schlapfer2014scaling}.
In recent years, increasing research attention is drawn to the complex challenges faced by modern cities, ranging from climate change to economic inequality, especially as the detailed data of social fabric is increasingly available~\cite{youn2016scaling}. Researchers are dedicated to understanding the complex and concerning phenomena of experienced segregation~\cite{nilforoshan2023human}, widening socioeconomic gaps~\cite{brelsford2017heterogeneity}, and prevalence of slums~\cite{brelsford2018toward}.

Motivated by these empirical studies, extensive previous works aim to design agent-based model to explain complex urban system from a bottom up perspective. For example, researchers propose to use diffusion limited aggregation model to reproduce fractal urban morphology as the process of physical particles~\cite{batty1989urban}. A later study uses correlated percolation model on spatial lattice to explain urban morphology~\cite{makse1995modelling}. Recent study finds the scaling laws of urban growth and fractal urban morphology can naturally emerge from agent-based model of human mobility~\cite{xu2021emergence}. In terms of complex urban challenges, agent-based model has been leveraged to reproduce and explain the ubiquitous segregation in urban space~\cite{schelling2006micromotives}. 
Researchers have also developed bottom up model to explain the emergence of varying levels of socioeconomic inequality as the provision of universal basic urban service change~\cite{brelsford2017heterogeneity}.

However, previous studies mainly use agents guided by simple rules. The recent advance of large language models provides unique opportunity to design generative agents with much more sophisticated intelligence. Such agents have been proved feasible in simulating virtue village~\cite{park2023generative} and company~\cite{hong2023metagpt}. 
By leveraging the human like intelligence and bias encoded in large language model, these generative agents have the potential to explain more complex social phenomena. For example, recent studies show large language model-driven agents exhibit similar content bias in transmission chain~\cite{acerbi2023large}, and can reproduce the typical social processes like \emph{wisdom of crowd}~\cite{aher2023using} and \emph{social conformity}~\cite{zhou2023sotopia}. However, these agents are all simulated in simple and virtual environment. It is still unclear if they are sophisticated enough to interact with complex urban environment. Hence, it is an important research direction to expose language model-driven agents to the rich information collected from urban systems or generated by realistic simulators.

\subsection{Urban Computing and Intelligence}
As a core concept of urban intelligence, urban computing~\cite{zheng2014urban} is to solve the urban issues like traffic congestion, energy consumption, and air pollution with the help of computer science algorithms. In the early years, the research focused on the spatial-temporal data analytics and management~\cite{li2008mining, zheng2008understanding, zheng2009mining, zheng2010geolife, yuan2010t} and its application on urban planing~\cite{zheng2011urban, yuan2012discovering}, transportation~\cite{yuan2010t, lou2009map, wang2014travel} and environment~\cite{rana2010ear,devarakonda2013real}. With careful designed data processing and fusion technics in different fields, these works achieved pretty good results. In the recent five years, with the application of deep learning methods, urban computing achieved significant process on various prediction problems in the urban space, e.g., traffic prediction~\cite{wang2018will, guo2019attention, li2023dynamic} and mobility prediction~\cite{zhang2017deep, feng2018deepmove, yao2018deep} which are widely used in transportation, epidemic modeling and environment. STResNet~\cite{zhang2017deep} first introduces the convolution neural network into the crowd flow prediction problem to better model the spatial correlation between regions. DeepMove~\cite{feng2018deepmove} proposes to utilize the power of recurrent neural network and attention mechanism to capture the periodic pattern of individual mobility. ASTGCN~\cite{guo2019attention} applies graph neural network to the traffic network to capture the spatial-temporal correlations between road segments. 

While these data analytics and prediction methods of urban computing helped us better understand the urban space from different aspects, they are limited for tackling many real-life issues which require counterfactual inference and decision making ability. Thus, recent works further extend the concept of urban intelligence into the new fields like behavior simulation~\cite{[Mobility2],[Trajectory1],[Trajectory3]} and decision making~\cite{wei2018intellilight,wei2019colight,zheng2023spatial}. MoveSim~\cite{[Mobility2]} simulates the human movement behaviors and is applied in the epidemic modeling. Further, SAND~\cite{[Trajectory3]} develops a knowledge-driven framework to simulate the human activities with Maslow’s need. Intellilight~\cite{wei2018intellilight} first utilizes deep reinforcement learning to enable the intelligent traffic light control. Zheng et. al.~\cite{zheng2023spatial} propose an artificial intelligence urban-planning model to generate spatial plans for urban communities with the help of graph neural network and deep reinforcement learning.

However, despite the above advancements, these methods usually simplify the assumptions of specific real-life problems and fail to solve the complex urban system issues in practice. Recently, the rapid development of LLMs with incredible human-like cognitive abilities provide us new opportunities to solve these issues. The integration of LLMs into the urban intelligence will enable the more real-world applicable methods and frameworks for pattern discovery, dynamics prediction and decision making in the urban space from a systematic view.

\subsection{Digital Twin City} 
Digital Twin City (DTC) refers to the emulation of a city in a digital environment, enabling real-time sensing, analysis, and optimization of urban systems through the use of data-driven models~\cite{zhou2018logic}. DTC stands as a significant trend in smart city research, finding applications in urban planning, traffic management, environmental protection and disaster response. Recent years have witnessed substantial progress in various information technologies, significantly accelerating the development pace of DTC. On one hand, advancements in sensor devices have facilitated the collection of vast amounts of data from multiple sources, including aerospace satellites~\cite{yang2021urban}, aircraft and drones~\cite{mohd2018remote}, smartphones and mobile terminals~\cite{calabrese2014urban}, smart wearable devices~\cite{salamone2021wearable}, industrial and household monitoring equipment~\cite{kim2021systematic}, and wireless sensing devices~\cite{rashid2016applications}. A recent focus has been on crowd-sensing methods that leverage distributed smart devices within a crowd, exhibiting distinctive characteristics~\cite{capponi2019survey, guo2022crowd}, focusing on integrating and analyzing digital footprints left by large-scale crowds to establish reliable and semantic-rich representations of group behavior across spatial domains. On the other hand, the analysis and optimization of urban data heavily rely on machine learning algorithms. The complex spatio-temporal data of DTC present challenges for forecasting and decision-making, and the advent of AI~\cite{ghaderi2017deep} has significantly improved data processing efficiency.  Advancements like differentiable decision trees~\cite{silva2020optimization} or knowledge graphs~\cite{bi2023bridged, yuan2023practice} can yield more informed and contextually sound outcomes through utilizing inherent comprehension of domain intricacies. Deep reinforcement learning provides a direct path to model spatio-temporal data through trial-and-error learning with agents~\cite{yuan2022activity} for urban decision-making tasks including traffic signal~\cite{mannion2016experimental} or navigation~\cite{liu2020robot}.

It is important to acknowledge that the digital twin city technology to solve urban problems still faces several challenges and problems. Firstly, there is a strong need to enhance the capability to process the multi-source urban data to improve credibility and realism. Secondly, the processing and computation of large-scale data pose challenges for real-time updates and evolution, and how to process complex instructions and provide human-friendly outputs, making DTC more accessible to policymakers and urban planners, is crucial for scenario testing and decision-making processes. Our proposed LLM-empowered fundamental city platform aims to solve these problems to upgrades DTC into the practice of urban problem solving.

\subsection{Large Language Models and Agents} 
Large language models~\cite{zhao2023survey} such as ChatGPT~\cite{chatgpt},  LLaMA~\cite{touvron2023llama}, Alphca~\cite{alpaca}, and GLM~\cite{zeng2023glm}, are the recent advances of artificial intelligence, which learns from the large corpus, with emergent abilities in understanding and generating language texts.
Since language is the most basic tool for humans interacting with the world~\cite{hauser2002faculty}, the human-like language ability endows large language models with high-level capacity, including reasoning and decision-making. Therefore, large language models are considered a promising approach for artificial general intelligence~\cite{bubeck2023sparks}.

To achieve urban generative intelligence, it is required to endow the existing large language with essential abilities for urban scenarios.
It is worth mentioning that there are some recent attempts to build city-related large language models.
Deng~\textit{et al.}~\cite{deng2023learning} proposed to fine-tune Llama-7B with Geoscience Academic Knowledge Graph and relevant research papers in the geoscience field. The fine-tuned model obtained the ability to understand these professional concepts, and support the basic QA tasks.
Similar solutions with fine-tuned large language models with geo-related corpus include GeoLM~\cite{li2023geolm} and GeoLLM~\cite{manvi2023geollm}.
Zhang~\textit{et al.}~\cite{zhang2023trafficgpt} utilize the large language model to help process the user query in traffic-related tools.
GeoGPT~\cite{zhang2023geogpt} took advantage of the ability in tool usage, considering large language model a bridge in connecting the practitioners with GIS software.
Despite these efforts, they are still limiting in only understanding some city-related concepts, without fully considering what abilities a real human has in the urban scenarios, \textit{i.e.}, urban generative intelligence, limiting these works' application.
In this paper, we pay attention to the construct the foundational model, simulation environment, and embodied agents for the urban context.

Despite the astonishing performance in various tasks of natural language processing, the large language models can also be used as an agent~\cite{wang2023survey,xi2023rise}, which can act and behave like a real human, serving as a virtual agent for personalized purposes.
That is, the agent can be a digital twin in various scenarios, such as representing a real human to interact with other humans in social networks in Metaverse applications~\cite{lee2021all}. 
One of the most critical challenges is to extend the language ability to more dimensions of models, from environment perception and action execution.
On the one hand, the environment can be represented as textual descriptions~\cite{park2023generative}, which can be naturally perceived by the large language models; on the other hand, recent advances build the multi-modal ability by the alignment among different modals~\cite{huang2023chatgpt}, including textual, visual, etc.
As for action execution, it is widely acknowledged that the agent can leverage various tools well, which extend the action space to language into real-world actions, supporting various interactions between the large language model agent and the environment. That is, it endows the large language models with the ability of embodied perception, reasoning, and action, which is also known as embodied agent~\cite{wang2023voyager,driess2023palm,mu2023embodiedgpt,zhang2023building}.
Specifically, these works approach the embodied tasks in the real-world environment, such as navigating and controlling robots, and take advantage of the reasoning and decision-making abilities of large language models. Moreover, to ensure the agents can take embodied actions, these works connect the textual output with a tool or another action-execution module.

However, these works only consider simple environments such as a room or virtual game environment. In addition, the tasks are relatively simple.
Unlike these works, in this paper, we present a far more complex problem in building fundamentally embodied agents in the city environment, and the agents can have almost all kinds of embodied behaviors of real humankind.

\subsection{Agent-based Modeling and Simulation}

Agent-based modeling and simulation is an important and powerful approach to modeling complex systems, such as the city system, understanding, analyzing, explaining, and even predicting the dynamics of the systems~\cite{baudrillard1994simulacra}. 
Generally speaking, simulation can be divided into macrosimulation and microsimulation.
Macrosimulation refers to simulating the system from an aggregate or high level, which focuses on the trends and behaviors within a system or a population without focusing on the individual-level characteristics. Specifically, macrosimulation may deploy several equations to describe how the critical variables in the system affect each other.
However, it is quite challenging to formulate the equations since real-world systems are always very complex, which motivates the microsimulation, which is also known as agent-based simulation. On the other hand, microsimulation focuses on individual entities within a system.

In general, agent-based simulation aims to model the behavior of individual components or agents to understand their interactions and how they collectively contribute to the overall system. For example, the famous Cellular Automata~\cite{wolfram1983statistical} is comprised of discrete cells, each following a set of rules based on their neighboring cells. 
The simulations based on setting rules for each individual can often showcase emergent behaviors, where complex patterns arise from simple interactions between individual components. 
Since it is general, agent-based simulation is extensively used in various fields, such as biology~\cite{an2009agent}, ecology~\cite{mclane2011role}, sociology~\cite{macy2002factors}, etc., to model systems where individual entities influence collective behavior.

The early attempts at agent-based simulation~\cite{brock1998heterogeneous,tesfatsion2006handbook} used some simple rules or formulas to guide how each individual behaved when faced with environmental change, which is easy to implement but makes it hard to capture complex individual behaviors accurately. After that, with the development of neural networks, for those individual decision factors that the simple rules cannot well capture, the neural networks are leveraged~\cite{farmer2009economy,geanakoplos2010leverage}.
Furthermore, recent works tend to deploy reinforcement learning-based agents~\cite{zheng2022ai}, for which each individual's goal in the simulation is to maximize the reward.

However, these agents are limited since they are not autonomous and require human-defined goals or rules, which motivates the large language model-driven agents.
In this paper, we present the UGI system, one of the main goals of which is to deploy large language model-based agents to simulate the complex dynamics of the city and further support various applications such as decision-making, etc.
The large language model agents are the up-to-date solution for agent-based modeling and simulation for the complex city system.

\subsection{Metaverse}
The Metaverse epitomizes a collective virtual shared space, formed from the fusion of physical and digital realities, typically accessed through immersive technologies such as virtual reality (VR)~\cite{lee2021all} and augmented reality (AR)~\cite{ning2023survey}. This digital universe offers an array of interactive experiences, social interactions, and economic activities, presenting various research questions and avenues for advancement. At its philosophical core, the Metaverse is entwined with the reality-virtuality continuum~\cite{chen2023philosophy}. On the reality aspect, it draws inspiration from the concept of the digital twin, meticulously replicating the physical world within the virtual sphere. This comprehensive replication captures physical objects, interactions, and dynamics, integrating reality into the digital landscape~\cite{fei2022digital}. Conversely, the virtuality facet of the Metaverse revolves around generating entities within the digital realm~\cite{binkley1997vitality}. These virtual creations, born from human imagination and innovation, surpass physical limitations and showcase human ingenuity in the virtual domain. Recent strides in Artificial Intelligence Generated Content (AIGC) notably advance this field~\cite{lv2023generative, qin2023empowering}.

Therefore, it comes as no surprise that research in this field is currently focused on two primary directions, one of which involves progress related to devices closely associated with the physical world. In contemporary times, Metaverse applications vary based on execution devices, such as tabletops, projectors, hand-held touchscreen devices, and headsets~\cite{goh20193d, quest2022oculus, apple2023vision}. These devices play a pivotal role in creating seamless, immersive experiences. Recent studies have delved into innovative solutions aiming to prevent information overload~\cite{dincelli2022immersive}, alleviate cognitive load~\cite{kim2019foveated}, explore eye-tracking technologies~\cite{robert2023analysing}, synchronize visual-motor responses~\cite{kondo2018illusory}, or leverage natural finger positioning~\cite{yamada2023one}. These efforts aim to create more accessible and intuitive entry points into the physical world. In the virtual realm, advancements in Artificial Intelligence (AI) have revolutionized artistic creation. With the success of diffusion models~\cite{rombach2022high} and applications like Midjourney, demonstrating high-quality, real-life image generation capabilities, creators within the Metaverse can now generate diverse types, contents, and styles of artworks through simple textual prompts. Similarly, generative works in film~\cite{esser2023structure}, audio~\cite{jin2022metamgc}, or poetry~\cite{liu2023research} have produced impressive content, ultimately providing users with experiences derived from reality but elevated beyond it. This revolution extends further to architectural designs~\cite{huang2018architectural}, landscape design~\cite{ardhianto2023generative}, and urban planning~\cite{zheng2023spatial}, offering exciting possibilities to create a metaverse city.

Looking ahead, the Metaverse is poised for rapid evolution and expansion, particularly in encompassing entire cities and creating an equitably accessible City Metaverse. Future developments may involve enhanced interoperability between diverse virtual environments, integration of AI for personalized experiences, advancements in haptic feedback systems~\cite{cui2022evaluating}, and exploration of blockchain~\cite{10.1145/3474355} for decentralized virtual economies. Our proposed foundational platform aims to build the future city metaverse with embodied agent to achieve urban generative intelligence.
\section{Architecture}

\begin{figure}[t]
\centering
\includegraphics[width=0.75\textwidth]{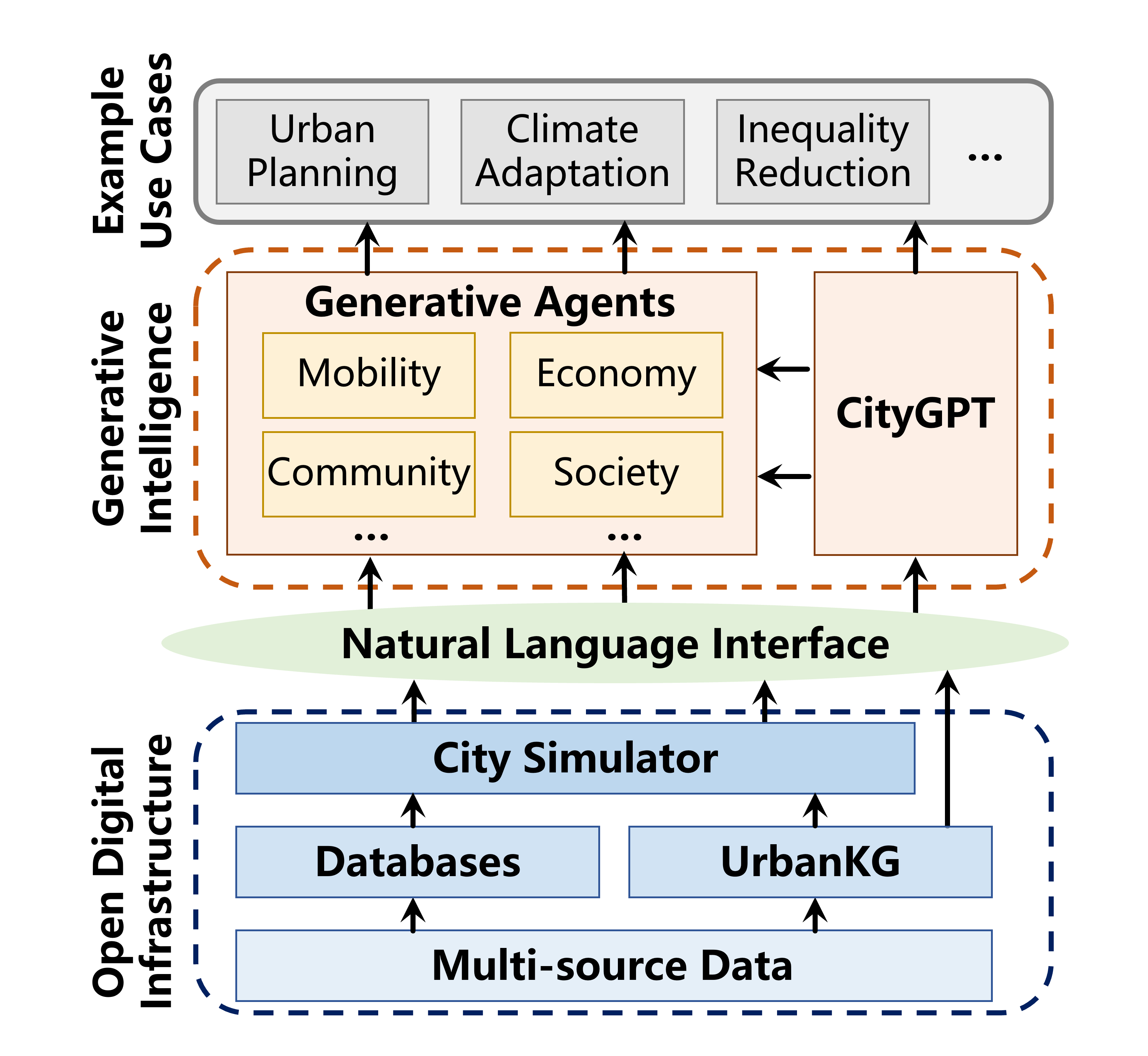}
\caption{Architecture of the foundational platform for urban generative intelligence.}
\label{fig1}
\end{figure}

Here, we present the overall architecture of our proposed Urban Generative Intelligence (UGI) platform (see Figure~\ref{fig1}). The key idea is to assemble the powerful city simulator, urban knowledge graphs (Urban KG) and various data streams as an open digital infrastructure. More importantly, the infrastructure will provide a standard language interface that enables the easy plugin of large language models and generative agents. It allows the generative intelligent models to conveniently access the computation power and factual knowledge in digital infrastructure, test strategies in various simulated scenarios and learn to evolve based on the feedback. Consequently, these empowered generative models will facilitate various downstream urban applications.
The key components of the presented architecture are elaborated as follows:

\textbf{Open Digital Infrastructure}: This component aims to provide a backbone system that integrates the data science resources and computational tools designed for urban problems. Specifically, it accesses various data streams in urban systems to collect massive spatial-temporal data of empirical urban activities, covering the aspects of spatial layout (e.g. points-of-interest and areas-of-interest), infrastructure distribution (e.g. road network and subway network), human behaviour (e.g. individual movements and collective mobility flows) and urban dynamics (e.g. traffic jams). These rich datasets are fed into a powerful city simulator, \emph{Mirage}~\cite{zhang2022mirage}, which can simulate the complex interactions between human, thing and space in an efficient and extensible manner. On top of the empirical observational data, this module can simulate various hypothetical scenarios efficiently, providing diverse environment to host intelligent agents. Besides, \emph{UrbanKG} module~\cite{liu2023urbankg} fuses various data streams and extracts factual knowledge, such as the spatial relation like ``border by'' and semantic relation like ``category of''.  
Urban KG provides the functions of construction, storage, and basic operations and algorithms of factual knowledge, which can facilitate easy access in generative intelligence models.

\textbf{Language Interface}: We design a standardized language interface to fully release the power of open digital infrastructure. City simulator, Urban KG and diverse data sources used to be difficult to access. They often require customized algorithms to configure city simulator, retrieve factual knowledge from Urban KG and integrate various data sources. Such obstacles limit their application in downstream tasks, and make their power inaccessible to advanced AI models. In our architecture, we design a user-friendly language interface to fully unleash the potential of the open digital infrastructure. It uses predefined natural language protocol to allow large language models and generative agents to conveniently leverage the computation power of city simulator and access factual knowledge from Urban KG. The standardized language interface reduces the barrier of developing language model-driven agents on top of the open digital infrastructure, which hopefully will foster the proliferation of generative urban agents.           

\textbf{Generative Intelligence}: On top of the language interface, we propose to train a foundation model customized for urban problems, \emph{CityGPT}. Specifically, CityGPT is a pre-trained large language model that encodes local urban knowledge via the language interface. It effectively leverages the reasoning capability and common sense in large language model, and greatly reinforced for specialized local urban problems. Empowered by this powerful city foundation model, we will design a series of generative agents in the dimensions of urban mobility, economy, community and society. These agents not only are capable of high quality decision making in various scenarios, but also can enable realistic agent-based simulations. Such agents combined with \emph{CityGPT} will release the power of generative intelligence to solve various important urban problems, such as urban planning, climate adaptation, inequality reduction, etc.

\section{Open Digital Infrastructure}

\begin{figure}[t]
\centering
\includegraphics[width=1\textwidth]{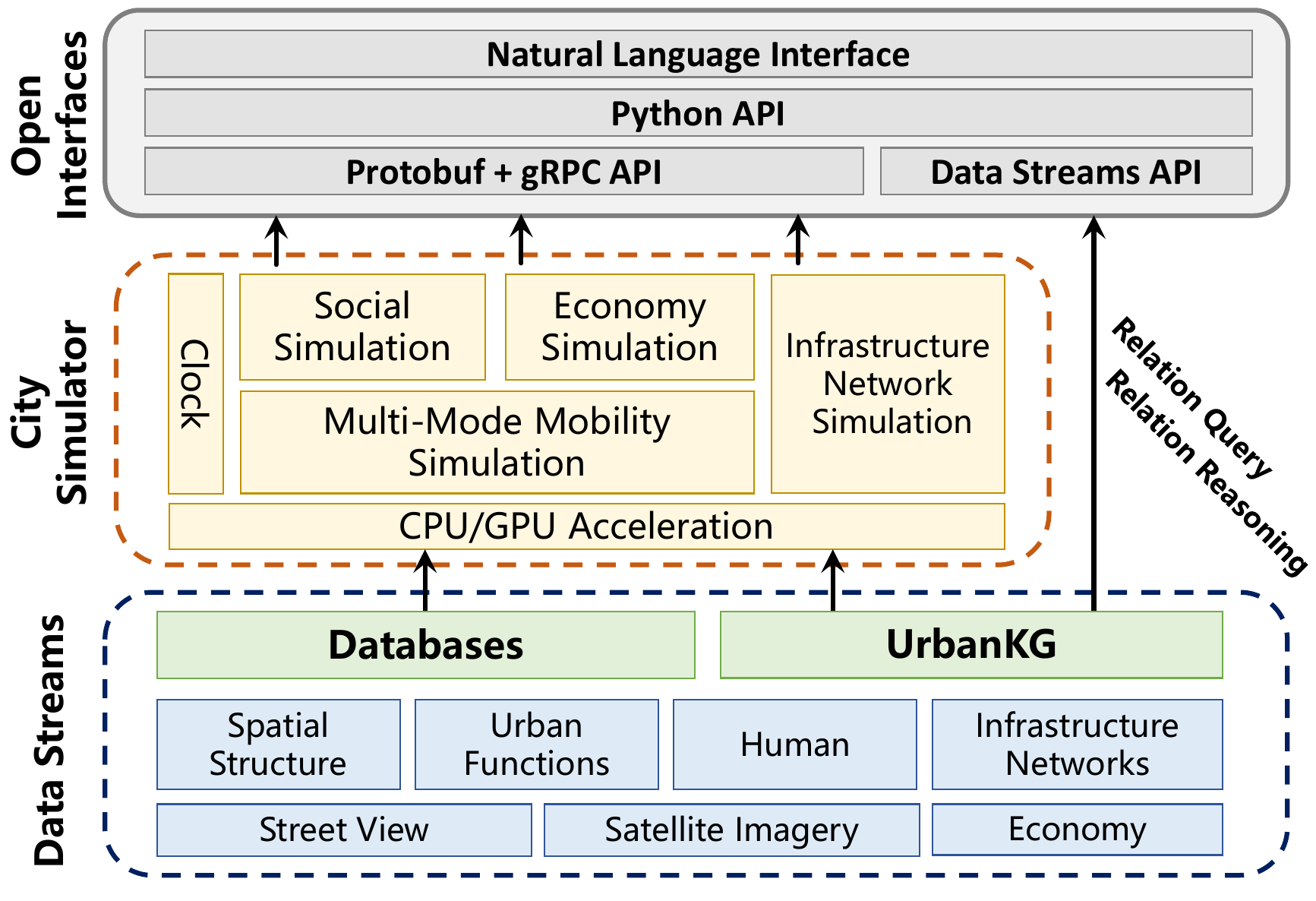}
\caption{The framework of the open digital infrastructure.}
\label{fig:infra}
\end{figure}

One question that must be answered on the road to building urban generative intelligence is how to create a real urban digital environment.
With a real urban digital environment, agents can use LLMs' capabilities to perform realistic interaction behaviors in this environment.
Therefore, they can address specific urban issues, simulate complex urban systems, and even assess the level of urban intelligence.
To build open digital infrastructure that satisfy the interaction requirements of agents, we start with urban modeling in Section~\ref{sec:urban_modeling}, build data streams of real urban data based on multiple sources in Section~\ref{sec:data_streams}, implement a computational engine (i.e. city simulator) that supports mobility, social, and economic simulations in Section~\ref{sec:citysimulator}, and finally provide open application programming interface (API) and natural language interface for agents and human users in Section~\ref{sec:interfaces}.
The whole framework of the open digital infrastructure is shown in Figure~\ref{fig:infra}.

\subsection{Urban Modeling}\label{sec:urban_modeling}

From the perspective of urban spatial structure, cities can be considered to be made up of many areas and a road network connecting each area.
These areas include commercial land such as shopping centers, industrial land such as factories, residential land such as neighborhoods, and public service land such as parks and sport fields.
In the work, we define these areas as areas-of-interest (AOIs).

From the perspective of urban functions, cities are represented as a collection of points-of-interest (POIs).
Restaurants cater to the dietary needs of the public, hotels provide temporary accommodation for travelers, and auto repair stores help people with vehicle problems.

To model the external and internal characteristics of the city at the same time, our urban modeling takes into account both the urban spatial structure and urban functions, including the urban road network, AOIs, and POIs.
In more detail, the urban road network contains two types of elements, roads and junctions.

After modeling the spatial structure and function of the city, we then consider the most critical element that makes up the city, human.
Under such modeling, human activity in the city is in terms of spatial structure moving between the urban road network and AOIs, while in the functional sense it is expressed as visiting different POIs at different times.
Human social behavior is then viewed as peer-to-peer and peer-to-cluster messaging.
Relationships in pairs or groups can be predefined through social networks (i.e. online socialization) or obtained based on spatial proximity (i.e. offline socialization).

To model the necessities that people need to live in cities, we also need to model infrastructure networks and economic systems.
Infrastructure networks, including power grids, water supply networks, communication networks, etc., are modeled as a topology with AOIs as vertices, and edges represent infrastructure conduits like electrical wiring, water pipes, etc.
Modeling of economic systems includes companies, individuals, governments, banks.
Between these entities, we model the basic economic behaviors of consumption, wages, taxes, and interest.

Through the above modeling, we are able to obtain a comprehensive description of people's lives in the city, including people's mobility, people's socialization, people's economic behavior, etc.
This will guide us in refining the construction of real data streams and the implementation of the city simulator.

\subsection{Data Streams}\label{sec:data_streams}

\subsubsection{Databases}

To model cities realistically, it is important to continuously collect real-world data and enrich the attributes of urban elements based on the data and the data preprocessing processes.
To achieve this, we build a pipeline that incorporates multiple data sources such as open source crowd-sourced data, research results~\cite{gong2020mapping,tatem2017worldpop}, and Internet application services to profile a city.
After preprocessing is complete, the city modeling data is stored in multiple databases to be used as input to the city simulator or available for call by the agent via the API.

For the urban spatial structure, i.e., the urban road network and AOIs, we first obtain the original urban road vertex-edge topology and AOI polygon boundaries from OpenStreetMap~\footnote{\url{https://www.openstreetmap.org/}}.
To improve the quality of the OpenStreetMap data to meet the needs of the city simulator, we perform a series of processes on the raw geometric data.
For the urban road vertex-edge topology, we first aggregate redundant vertices and edges so that each vertex corresponds to a real-world junction and an edge corresponds to a real-world road.
For the edges, we assign the number of lanes and the speed limit to them based on the road class provided by OpenStreetMap.
Then, we identify the junction morphology based on the topology and the geometry of the edges, distinguishing between ramps, crossroads, T-intersections, etc.
Based on the junction morphology, we establish the road to road connectivity at the junction and assign their turn types and signal phases.
Finally, we can obtain an urban road network with the following elements and attributes:
\begin{itemize}
    \item \textbf{Road:} ID, road's geometry, number of lanes and speed limit.
    \item \textbf{Junction:} ID, list of road IDs connected to the junction, connectivity between roads with the number of lanes and the turn types, and signal phases.
\end{itemize}
For the AOI data, we use only the polygon boundary data provided by OpenStreetMap.
To enrich the attributes of AOIs, land use~\cite{gong2020mapping} and population data~\cite{tatem2017worldpop} are matched to AOI polygons and used to add the type attribute and the population size attribute to the AOI.
Besides, we also match the AOI polygons with close roads to get the spatial locations of the connection points between the AOI and the neighboring roads, and establish the topological association between the AOI and the urban road network.
Thus, we can obtain AOIs with the following attributes:
\begin{itemize}
    \item \textbf{AOI:} ID, boundary geometry, land use type, population, and road connection points.
\end{itemize}

For POI data representing urban functions, we use UrbanKG's data sources to benefit from the spatio-temporal correlation knowledge and data fusion of urban domains provided by UrbanKG.
We match the POI data to the AOI containing it based on its latitude and longitude coordinates.
In this way, elements describing urban functions are linked to the spatial structure of the city through spatial subordination, which helps to understand and control the intentions (e.g. go for leisure) and actions (e.g. drive to the park) of human activities in cities.
We can obtain POIs with the following attributes:
\begin{itemize}
    \item \textbf{POI:} ID, coordinate, name, category, and belonging AOI ID.
\end{itemize}

It is very hard to get direct access to the activities of all people in the city.
However, through data analysis and model training on sampled human travel records (e.g., check-in sequences, GPS tracks, etc.), deep learning models~\cite{rong2023interdisciplinary,rong2023complexity,yuan2022activity,yuan2023learning} can help us generate and restore the full amount of human mobility behavior in the city that conforms to the real pattern.
The generated human activity contains the following attributes:
\begin{itemize}
    \item \textbf{Person:} ID, home position, list of trips.
\end{itemize}
A trip is characterized by a tuple $(P_{s}, P_{e}, t_{s}, mode)$, where $P_{s}$ and $P_{e}$ denote the starting and ending positions of the trip, $t_{s}$ represents the starting time of the trip, and $mode$ indicates the mobility mode used during the trip, such as walking, driving, or biking.

For the infrastructure network, we use heuristics for construction.
The methods consider each AOI as a vertex, calculate the resource demand based on the regional population and land use type, and then obtain the higher-level vertices and the edges based on the aggregation of the resource demand.
Finally, we can obtain an infrastructure network topology with a hierarchical structure and use it for infrastructure network simulation.
In general, infrastructure networks have the following attributes:
\begin{itemize}
    \item \textbf{Vertex:} ID, coordinate, level and belonging AOI ID.
    \item \textbf{Edge:}  vertex pair, line geometry and level.
\end{itemize}
For different infrastructure networks, specific fields are also added to match the corresponding simulation program inputs.

The city's economic performance data are better publicized.
By crawling the public information from internet platforms such as enterprise information disclosure platforms, recruitment platforms and real estate agency platforms, we get the information of enterprises in the city as well as the main industries, wage levels, consumption levels and rent levels in different areas.
This data is eventually matched to the AOI by adding the following fields to the AOI:
\begin{itemize}
    \item \textbf{Enterprises}: name, category, registered capital, number of employees and average wage.
    \item \textbf{Consumption}: per capita consumption of different POI categories.
    \item \textbf{Rent}: average rent.
\end{itemize}

In order to add more real city information and build a multimodal data base, we finally introduced images as an additional data source.
These images include satellite images and street view data.
Satellite image data is organized in tiles and the data source is from Mapbox~\footnote{\url{https://docs.mapbox.com/api/maps/raster-tiles/}}.
Street view data is obtained from Baidu Maps~\footnote{\url{https://map.baidu.com/}}, and the spatial coordinates where the images are located are spaced at intervals of 100 meters from each other.

\subsubsection{Knowledge Graphs}

At the urban functional level, the complex relationships between POIs  further constitute the uniqueness of the city.
However, it is difficult to process and mine the correlation between POIs and aggregate and distill massive data based solely on the attributes of POIs.

In the field of data mining, knowledge graphs~\cite{wang2017knowledge,hogan2021knowledge} are a means of effectively organizing massive data and knowledge, which can help users quickly retrieve other entities that are related to a given entity.
Inspired by the idea, UrbanKG~\cite{liu2021knowledge} is proposed to build the relationship network of entities within the city.
UrbanKG builds a variety of relations for urban entities.
Of these, the relations on the urban spatial structure include \verb|borderBy|, \verb|nearBy| and \verb|locateAt|, \verb|belongTo|.
The relations regarding urban functions include \verb|brandOf|, \verb|cate1Of|, \verb|cate2Of|, \verb|cate3Of|, \verb|competitive|, \verb|coCheckin|, \verb|similarFunc|, \verb|provideService|, etc.

Besides, for multi-modal data, UrbanKG establishes two relationships \verb|satelliteImageOf| and \verb|streetViewOf|.
These relational categories designed by expert knowledge and the data-driven fact set provided by UrbanKG enhance the understanding of urban spatial structure and urban functions and also provide more effective information input to agents.

\subsection{City Simulator}\label{sec:citysimulator}

The open digital infrastructure computational engine is called the city simulator.

Firstly, the city simulator~\cite{zhang2023city,zhang2022mirage} can efficiently simulate the interactions between human and urban space in large-scale cities.
In detail, it simulate the movement of agents in the urban space between roads and AOIs using multiple mobility modes (e.g. driving, walking, etc.), and provide agents with the ability to sense the environment like obtaining its current road or querying specific AOIs' and POIs' information by IDs.
The agent's mobility task in the simulator is described as a list of trips.
Therefore, the agent's behavior can be controlled by modifying its trip list.
Overall, city simulators provide agents with sense and control.
For example, the agent can use the city simulator's interface to obtain the current road, query the AOI and POI information on the road, and then control the agent to walk to the AOI where a restaurant type POI is located.

As the most important part of open digital infrastructure, the city simulator need to be able to host the access of a variety of generative agents and provide excellent computing efficiency to ensure the speed of city-level simulations.
To this end, the city simulator has implemented numerous software design and optimization for urban simulation, and achieves the following features:

\begin{itemize}
    \item \textbf{Multiple Modes Simulation:}
    The city simulator supports the simulation of multiple mobility modes, including driving, walking, biking and public transportation.
    This comprehensively models how people commonly move through space in cities.
    In order to realistically simulate human mobility behavior, we adopt the widely used IDM car-following model~\cite{treiber2000congested} and MOBIL lane-changing model~\cite{kesting2007general} to simulate driving, and the PCS~\cite{zhang2022physics} model to simulate walking.
    \item \textbf{Clock Synchronization:}
    During the simulation, the city simulator and the accessed agents must be able to keep the simulation time synchronized.
    In order to achieve this goal, the clock synchronization mechanism introduced in the Mirage framework~\cite{zhang2022mirage} is embedded in the city simulator, and the communication interface is opened as a necessary link for agent access.
    \item \textbf{Computing Acceleration:}
    Through reasonable design of control flow and data flow, and the introduction of an efficient indexing subsystem~\cite{zhang2023city}, the city simulator achieves computing acceleration of more than 10 times compared to wall clock time for nearly one million agents at the urban scale.
    This can help the agent quickly explore, learn and evolve.
    \item \textbf{Various Data Retrieval Interfaces:}
    The city simulator models the spatial structure of the city.
    Therefore, it also provides a rich data retrieval interfaces about the spatial structure and the agents running in the spatial structure.
    These retrieval interfaces include retrieval of urban spatial topological structures (e.g. which roads the specified AOI is connected to) and runtime status (e.g. how many people are in the specified AOI now).
    These retrieval interfaces provide two access forms: pull and push.
    The specific details will be introduced in Section~\ref{sec:api}.
    \item \textbf{Unified Control Interface:}
    In order to simplify the control of the agent, the control interface of the agent is unified into modifying the agent's trip list.
    By modifying the trip list, the caller can control the agent's attributes such as stay time, departure time, destination, and mobility mode.
\end{itemize}
On top of the mobility simulation, we also add the ability to social message propagation and financial flow mechanism based on the extended friendliness of the city simulator.
The social message propagation mechanism allows a person to send a message to a specific list of people or broadcast it to the surrounding crowd, which enables online and offline socialization, respectively.
Financial flows are triggered primarily based on a person's visit to a POI.
Depending on the POI category, the city simulator models consumption, income and taxes.
The simulation of interest is triggered periodically.
Through these features, the city simulator can realistically model the life activities of agents in urban space, orderly access a large number of generative agents, and provide rich sense capabilities and a simple and unified control method.
This provides an environment for exploration, learning and evaluation for agents.
Agents can use the city simulator to enhance their understanding of the city, and even deduce the future evolution of the city and find optimal decisions.

Using the distributed architecture described in Mirage~\cite{zhang2022mirage}, the simulation of the infrastructure networks is implemented as multiple independent extensions in the city simulator.
We support to integrate PYPOWER\footnote{\url{https://github.com/rwl/PYPOWER/}} as grid simulation, WNTR~\cite{klise2018overview} and SWMM~\cite{gironas2010new} as water supply and drainage simulation, and the digital twin system for mobile networks~\cite{gong2023scalable} as communication simulation.

\subsection{Open Interfaces}\label{sec:interfaces}

\subsubsection{Application Programming Interface}\label{sec:api}

As the open digital infrastructure in UGI, its capabilities need to be exported through open interfaces.
For experienced developers, we provide two layers of application programming interface (API).
The first layer is based on Protobuf~\footnote{\url{https://protobuf.dev/programming-guides/proto3/}} and gRPC\footnote{\url{https://grpc.io/}}.
We use Protobuf to standardize data structures including roads, junctions, AOIs, and POIs.
gRPC is used to implement communication between the caller and the city simulator to achieve clock synchronization, data retrieval, and control.
Actually, the gRPC implementation we use is Connect~\footnote{\url{https://connectrpc.com/}}, which is a simple, reliable and interoperable library to provide both browser (i.e. JSON format message on HTTP) and gRPC-compatible APIs.
For users who do not want to see the underlying communication implementation, we provide a higher-level Python API.
Through the Python API, users can interact with the open infrastructure in the form of Python function calls and receive responses in Python's basic data format (e.g. dict and list), which is more familiar to researchers.

For the original data from the data streams, encapsulation of database access is provided in the Python API.
Relation queries and reasoning to UrbanKG are also included in it.

Through these APIs, users can not only access the original data provided by the data streams, but also can \textbf{sense} the environment and \textbf{control} agents in the city simulator.
The main sense APIs are as follows:
\begin{itemize}
    \item \textbf{GetAoi}: get the runtime status of the specific AOI.
    \begin{itemize}
        \item \textbf{Input:} AOI ID.
        \item \textbf{Output:} list of people IDs, number of recent entries and departures.
    \end{itemize}
    \item \textbf{GetRoad}: get the runtime status of the specific road.
    \begin{itemize}
        \item \textbf{Input:} road ID.
        \item \textbf{Output:} list of vehicle IDs and pedestrian IDs, average speed and congestion level.
    \end{itemize}
    \item \textbf{GetPerson}: get the runtime status of the specific person.
    \begin{itemize}
        \item \textbf{Input:} person ID.
        \item \textbf{Output:} coordinate, speed, direction, trip currently in progress.
    \end{itemize}
\end{itemize}
The unique control API is as follows:
\begin{itemize}
    \item \textbf{SetTrips}: modify a person's trip list to change the person's moving target, departure time, and mobility mode.
    \begin{itemize}
        \item \textbf{Input:} person ID and new list of trips.
        \item \textbf{Output:} None.
    \end{itemize}
\end{itemize}

In particular, in order to avoid excessive processing pressure caused by polling, the city simulator provides a push mechanism for the retrieval of runtime environmental information.
The client can specify to monitor the changes of a certain element through the API.
When the element changes, the city simulator will push a message to the client to trigger the corresponding processing logic.
The list of triggers that support the push mechanism includes, but is not limited to, the following scenarios:
\begin{itemize}
    \item Someone enters the specific AOI.
    \item Someone leaves the specific AOI.
    \item Someone enters the specific road.
    \item Someone leaves the specific road.
    \item The specific person starts a trip.
    \item The specific person finishes a trip.
\end{itemize}
For example, the client can monitor the entry of a person into the specified AOI.
When someone enters the AOI, the client will receive the information about the person who enters the AOI, so that it can control the future behavior of the person.

\subsubsection{Natural Language Interface}

For users without programming skills, we also provide a natural language interface.
The natural language interface is a further encapsulation of the API.
Users can use some standardized natural language instructions to complete functions such as data retrieval and agent control like ALFWorld~\cite{ALFWorld20}.

For instance, in the natural language interface, the information to retrieve AOI is expressed in the following form:
\begin{itemize}
    \item \textbf{Request:}
    Get AOI with ID 500000000.
    \item \textbf{Response:}
    The AOI with ID 500000000 has an area of 26059 square meters, a population of 1219, the land use type is commercial land, contains 51 POIs, and is connected to roads 10, 11 and 23.
\end{itemize}
The control of the agent is expressed in the following form:
\begin{itemize}
    \item \textbf{Request:}
    Set agent with ID 1000 to drive to AOI 500000001 at 09:20, and then walk to AOI 500000010 at 11:00.
    \item \textbf{Response:}
    OK.
\end{itemize}

\section{Foundation Model and Agent}

\subsection{CityGPT: Large Language Model for Urban Generative Intelligence}

\begin{figure}[t]
\centering
\includegraphics[width=0.9\textwidth]{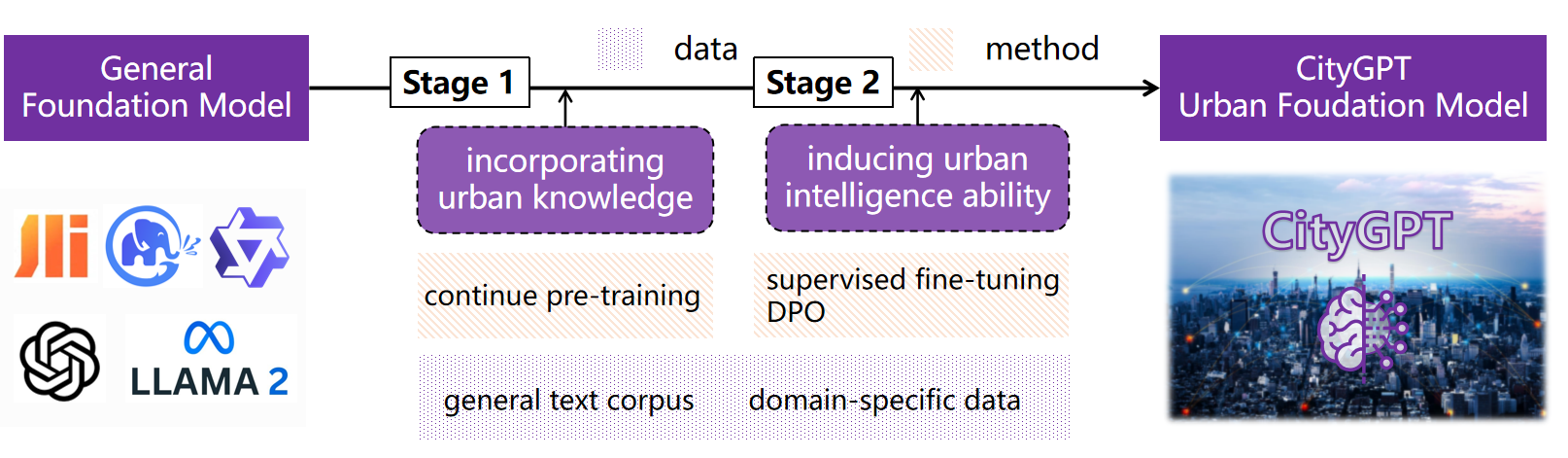}
\caption{Training procedure of CityGPT.}
\label{fig:citygpt}
\end{figure}

As the fundamental component of platform, the large language model plays the critical role of empowering the city agents with general and specific skills in the urban generative intelligence. In other words, the ability of the large language model determines the upper limit of whole system. There exists many open source large language model~\cite{yang2023baichuan,touvron2023llama,bai2023qwen}, which are build for general purpose. While these models perform well on common tasks like daily chat and text generation, they perform inefficiently even fail to support the generative city agent in the urban space in many cases due to the lack of domain-specific background knowledge~\cite{wang2023robustness} and related skills. Thus, we need to enhance the general large language model to meet the requirements of city agents modelling in the urban space. The whole enhancement procedure is presented in the Figure~\ref{fig:citygpt}. We introduce two steps to refine the general large language model to obtain the CityGPT. Firstly, we aim to incorporating the multi-source urban knowledge into the general large language model. Then, based on the output of the first step, we design sufficient methods and training datasets to induce the related skills of it for urban intelligence. 

\begin{figure}[t]
\centering
\includegraphics[width=1.0\textwidth]{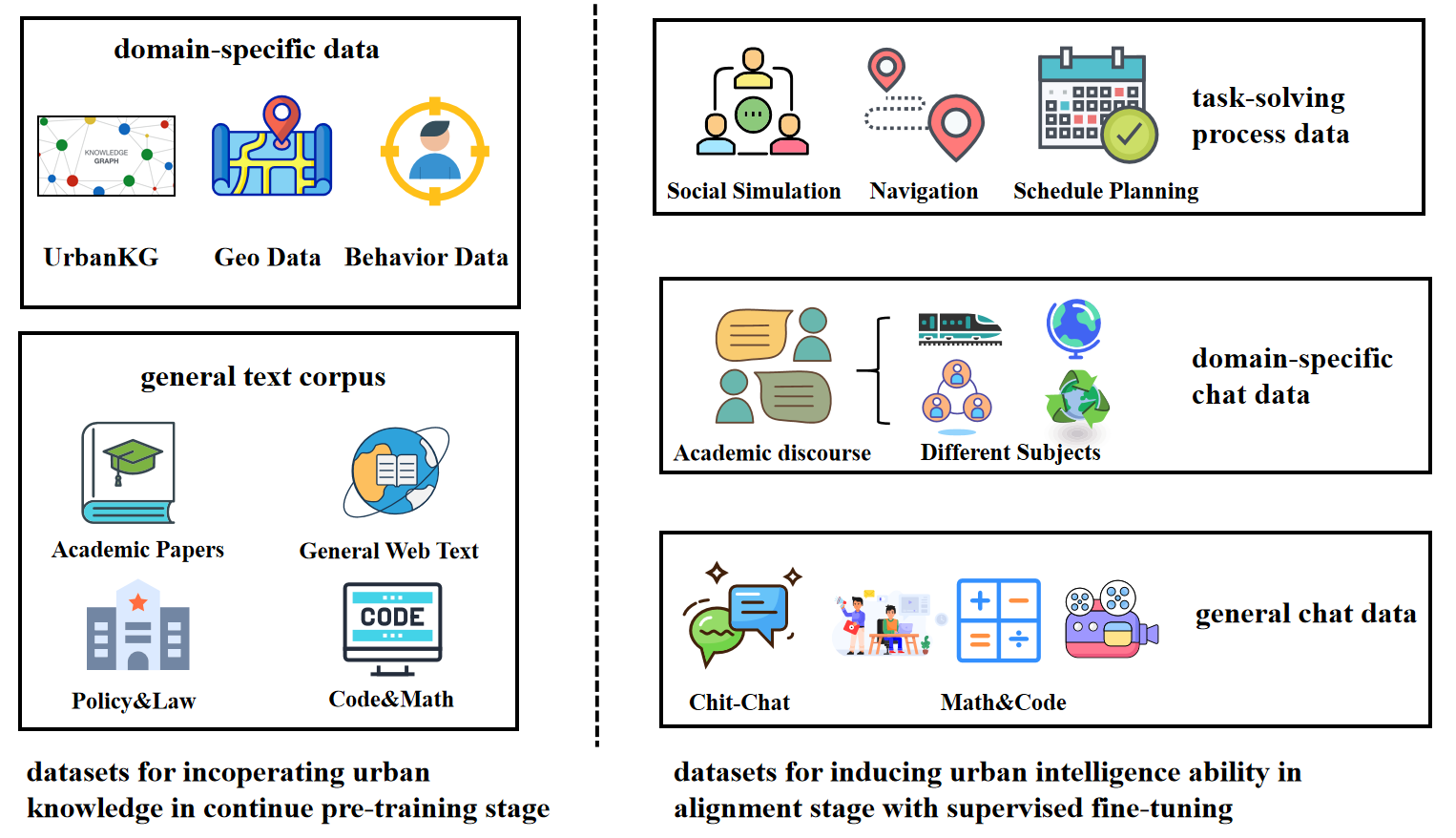}
\caption{Training data of CityGPT.}
\label{fig:data}
\end{figure}

\textbf{Stage 1: incorporating the urban knowledge.} As claimed in open source models~\cite{yang2023baichuan, touvron2023llama, bai2023qwen}, most of the training corpus are from the public web text like common crawl project and only limited common data processing methods are applied due to the high-cost of handling the large volume various data. However, domain specific data like urban knowledge data are usually not open and specific data processing method is fundamentally necessary for utilizing them. Thus, incorporating the carefully processed urban knowledge into the general large language model becomes an essential step. There are several methods for incorporating knowledge into the large language model, including retrieval based methods~\cite{lewis2020retrieval}, fine-tuning based methods~\cite{fu2023revisiting}, and continue pre-training methods~\cite{ke2022continual}. The retrieval based methods rely on the effective retrieval mechanisms and only limited knowledge can be utilized for each single use. While fine-tuning methods can introduce more knowledge than the first one, their capacity are still limited and may amplify the hallucination of large language model when requiring model to output things which not learned during its pre-training. Thus, we apply continue pre-training methods to incorporating urban knowledge into the general large language model. 

As shown in the left part in the Figure~\ref{fig:data}, we collect general text corpus (e.g., domain-specific research papers, high-quality codes and online web text) and domain-specific data (e.g., urban knowledge graph~\cite{liu2023urbankg}, geographic data, human behavior data~\cite{feng2018deepmove}) to conduct the continue pre-training process. We follow the data cleaning rules in open source LLMs~\cite{yang2023baichuan, bai2023qwen} to process the general text corpus data. As for the domain-specific data, we process them case by case. For knowledge graph data, we apply the universal knowledge-text prediction task~\cite{sun2021ernie} to integrate knowledge into model. Given a pair of tuple from urban knowledge graph, we use ChatGPT to generate a related text sentence. Different from the original paper, we combine the tuple and the related text as a training instance in the pre-training. For geographic data, we directly use the widely-used geojson~\cite{butler2016geojson} format in the GIS community to organize each object as a single training instance. For human behavior data, we alignment their elements with the aforementioned two data source and regard each behavior session as a single training instance. We choose the open source Baichuan2-7B~\cite{yang2023baichuan} as the base model and utilize deepspeed~\cite{rasley2020deepspeed} to continue pre-train the model on a single machine with eight A100 GPUs for about one week. Other general large language model can also be selected and we take Baichuan2-7B as an experimental example. The continue pre-trained model is called as CityGPT-base model.

\textbf{Stage 2: inducing the urban intelligence ability.} After the continue pre-training procedure, the based model acquires various urban knowledge. While we can directly extract information from the base model via in-context learning method, it requires few-shot demonstrations during the use which are not easy to construct. Besides, we want CityGPT learn to response based on the injected factual urban knowledge and follow the instruction from agents with standard output formats, e.g., JSON. Thus, we need the alignment procedure~\cite{ouyang2022training} to induce the these general and specific ability of model. Follow the common practice of large language model~\cite{yang2023baichuan, touvron2023llama, bai2023qwen}, we apply supervised fine-tuning and DPO~\cite{rafailov2023direct} to achieve this goal. The dataset used in the supervised fine-tuning stage is shown in the right part of Figure~\ref{fig:data}. Specifically, we build a domain-specific alignment dataset with three kinds of data: general purpose chat datasets~\cite{platypus2023,xu2023wizardlm}, domain-specific chat datasets and task-solving (e.g., schedule planning, and navigation task in urban space) chat dataset. The open source general purpose chat datasets aim to teach model how to chat with people fluently. Similar to the self-instruct framework~\cite{wang2022self, ding2023enhancing}, we build the domain-specific chat dataset by generating question-answer pair with ChatGPT on the domain specific text corpus. Task-solving dataset is built by solving classic urban tasks with the assistance of ChatGPT and external tools. It is noted that while urban knowledge data like UrbanKG are directly added in the continue pre-training stage, we also use the UrbanKG as an external knowledge source when solving the specific tasks. Details about the task-solving process of different urban tasks can refer to the following sections. After supervised fine-tuning CityGPT on above datasets, we use UltraFeedback~\cite{cui2023ultrafeedback} with DPO to align it with human preferences. We use trl~\cite{vonwerra2022trl} with packing strategy to accelerate the training procedure.

After the above two-stage training procedure, we obtain the CityGPT, an urban knowledge enhanced foundation model for urban generative intelligence. In the following sections, with sufficient prompts as inputs, CityGPT will follow the instructions to complete different tasks.

\subsection{General Framework for Generative City Agents}

\begin{figure}[t]
\centering
\includegraphics[width=0.75\textwidth]{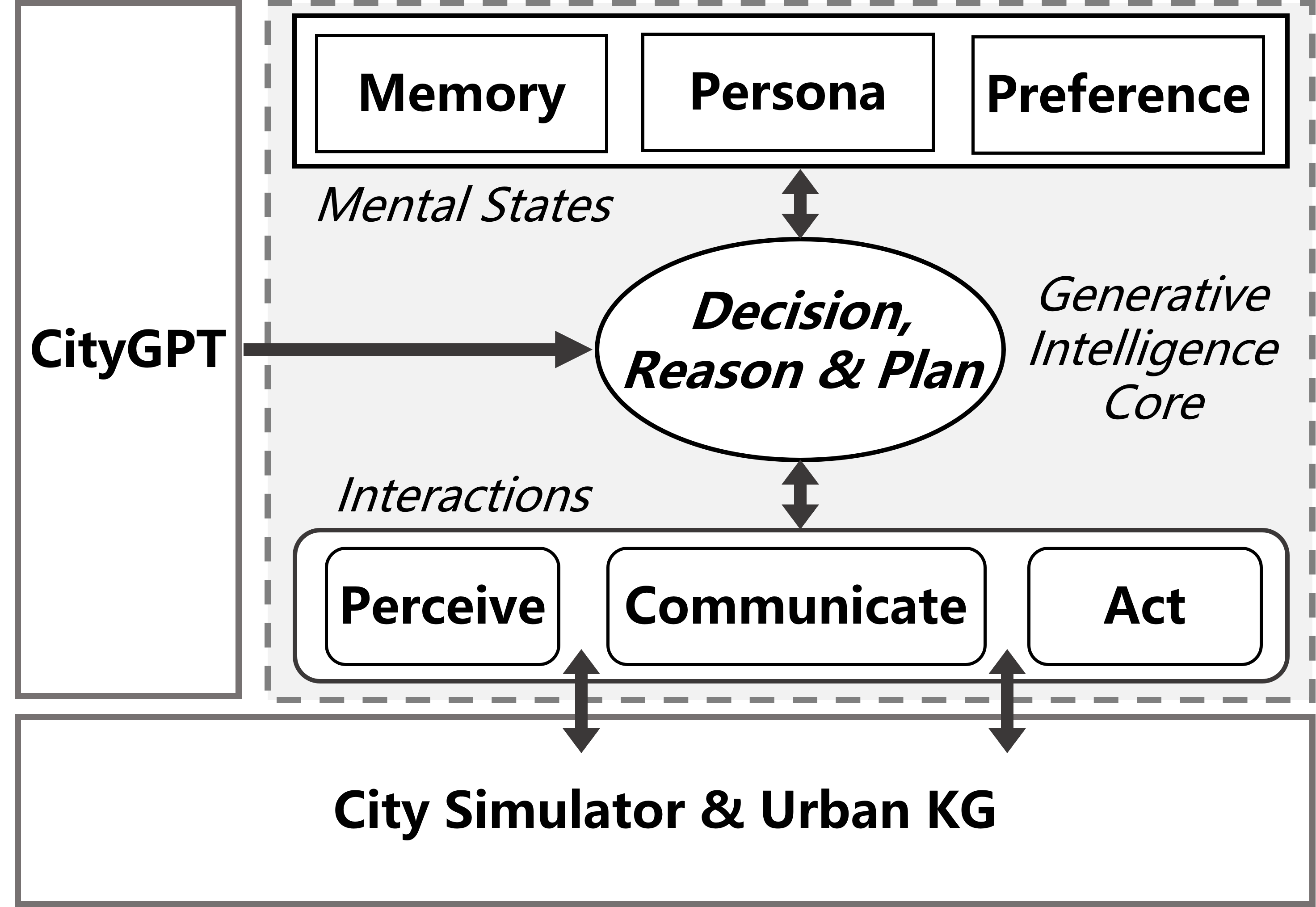}
\caption{A general framework for embodied generative agents in urban space.}
\label{fig2}
\end{figure}

Here, we present a general framework for embodied generative agents in urban space (see Figure~\ref{fig2}). This framework leverages generative foundation model as intelligence core, and it is built upon the open digital infrastructure of city simulator and UrbanKG knowledge graph. Following embodied cognition hypothesis~\cite{wilson2002six}, this framework allows generative agents to harness the realistic embodied feedback provided by the digital infrastructure and evolve its intelligence in simulated urban environment. 
The proposed framework aims to extract a unified conceptual abstract for most generative city agents, and provide enough flexibility for customization in various applications. Specifically, the autonomous agents under this framework have the \emph{Mental States} components of \emph{memory}, \emph{persona} and \emph{preference}. The \emph{memory} component stores the history of past behaivour and interactions, \emph{persona} component assign the agents a specific profile to leverage the role play capability of language model, and \emph{preference} component allows personalizing the agents with high-level language description. Besides, the \emph{Interaction} components include: \emph{perceive} module that senses the simulated urban environment; \emph{act} module that registers behaviours or status changes in city simulator; and \emph{communicate} module to exchange information with other agents. Finally, these agents use city foundation model as its generative intelligence core, which can comprehensively model the internal \emph{Mental States} and external \emph{Interactions} to generate appropriate behaviours. To better illustrate our framework, we provide several concrete agent designs as below, focusing on two major categories of urban problems, \emph{i.e.}, simulating urban phenomena and informing complex decision making.

\subsubsection{Simulation Agent: Generating Individual and Collective Behaviour}

Complex urban phenomena are driven by the spatiotemporal  agglomeration of micro activities in physical, economic and social domains. Understanding the underlying micro mechanisms and the emergence process plays an important role in modeling and managing urban systems, necessitating the simulation of complex urban phenomena with micro autonomous agents. Here, we present three design examples of embodied agents under the proposed general framework, which are customized for the simulation of the basic urban activities in physical, economic and social domains, respectively.

\textbf{Physical mobility}: 
This agent aims to simulate individual activities and movements within urban environments, with the objective of creating trajectories that mirror real-life patterns. In addition to generate logical individual mobility behavior, we also want to reproduce the statistical distribution of collective movements by simulating a population of these agents, such as reproducing the daily number of commuters between two locations. Traditional simulation models often use simplified rules to guide agent’s mobility behaviour, but they lack depth in understanding the rich semantic in urban mobility, such as the function of a specific place and characteristics of diverse demographic profile. The generative intelligence in city foundation model offers a promising alternative. It excels in common sense reasoning and has deep knowledge of the local environment. These features equip simulation agents with accurate prior of social norms and human behaviour patterns, contributing to more plausible simulation outcomes. Besides, the flexibility of language models, particularly through their prompt-based mechanism, enables more logical and realistic reasoning in behavior simulation.

We propose a generative agent that involves the \emph{memory}, \emph{persona}, \emph{perceive} and \emph{act} modules in the proposed general framework, combined with reasoning core driven by CityGPT. The \emph{memory} module serves as the knowledge base of historical mobility patterns, and \emph{persona} module holds demographic profiles of the simulated agents. The designed agent will generate mobility behaviour step by step based on its historic movements and the preference inferred from demographic profiles. Besides, it will also jointly consider the contextual information \emph{perceived} from simulated urban environment, such as current time and road traffic. After generating a mobility behaviour, it will register its new locations in simulator via \emph{act} module. We also design an anchor detection mechanism that will generate a routine for each agent based on its profile, such as the work schedule, serving as the anchor points of its daily life. The reasoning core is prompted to avoid violation with these anchor points during generation. Such agent designs allow the simulation of realistic, personalized and coherent urban mobility behaivours, which are the most essential micro urban activities in physical domains, reflecting the interactions between urban dwellers and the access of various urban resources.

\begin{figure}[t]
\centering
    \includegraphics[width=0.8\linewidth]{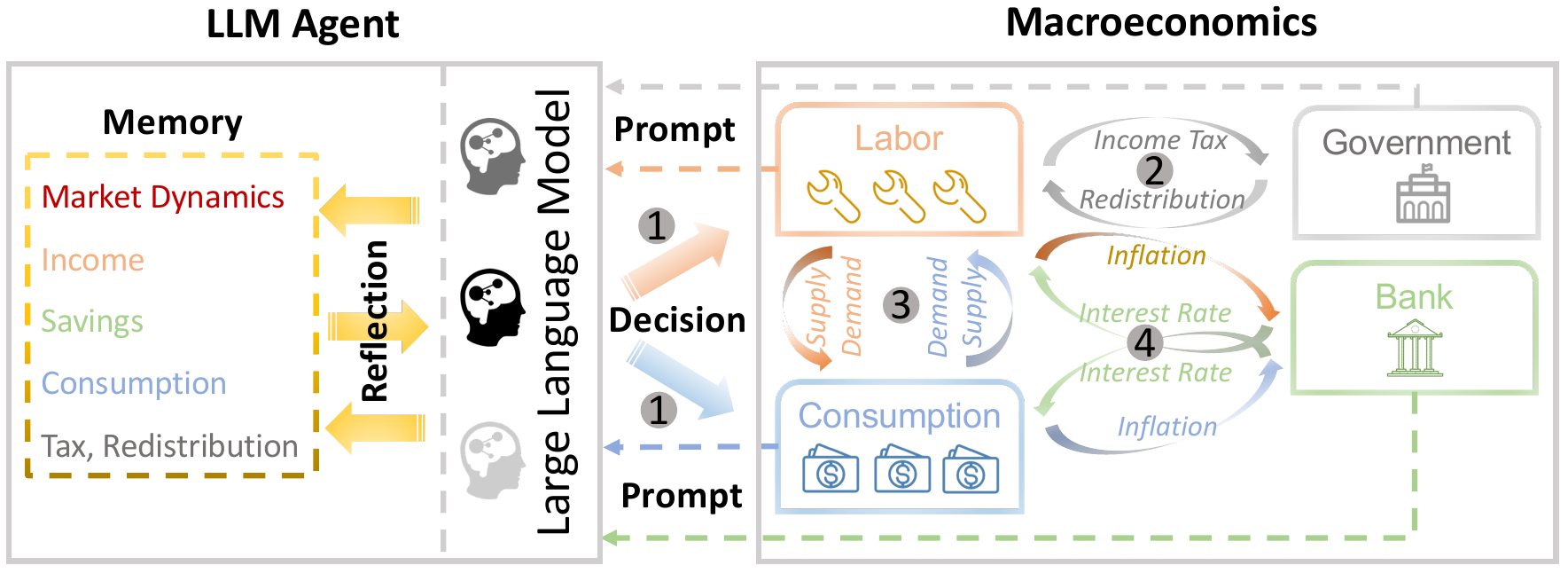}
\caption{The illustration of LLM-driven agents for economic simulation~\cite{li2023large}.}
\label{fig:econ-framework}
\end{figure}

\textbf{Economy activity}: 
For the economy, agent-based modeling and simulation is a promising solution for understanding and predicting the dynamics of economic systems~\cite{farmer2009economy}.
Specifically, when predicting economic indicators such as GDP, unemployment rate, etc., 
Traditional methods, such as econometrics~\cite{smets2003estimated}, cannot handle some complex real-world scenarios.
For example, for one of the most famous econometrics methods, Dynamic Stochastic General Equilibrium (DGSE)~\cite{hayashi2011econometrics}, sometimes there is no feasible solution for the equilibrium.
That is, the agent-based simulation can construct multiple heterogeneous agents to describe each user in the ecosystem and then define what kind of economic behaviors the agents can have. The major objective of the agent-based simulation in the economy is to observe both the emerging phenomenon from the perspective of macroeconomics and behavioral economics, which can regarded as an environment to support theory validation and decision-making.
To construct the economic agents, our previous work~\cite{li2023large} follows the well-acknowledged simulation mechanism illustrated in Figure~\ref{fig:econ-framework}. 

Specifically, there are four components: labor, consumption, financial markets, and government taxation, covering the primary components of existing macroeconomic simulations.
Specifically, the agent is deployed to simulate the two most critical decisions the real human will make in real life: going to work (earning money) and consumption (spending money).
The government agents decide the tax policy, and the bank agents adjust interest rates based on market inflation or deflation. 
From the macroscopic perspective, the system can observe the dynamics of overall labor and consumption markets. 

\textbf{Social interaction}: 
Human social behaviors can also be simulated with large language model-empowered agents. Specifically, the social agents require human-like abilities in social behaviors, \textit{i.e.}, interacting with other individuals in the city system.
For social activities, there are both online and offline social networks, \textit{i.e.}, the communication can occur in both online social networks or just via chatting in a room.
The social agent simulation mainly focuses on how information propagates on the social network and its further impact on the individuals.
That is, the LLM-driven agent can first shape their social awareness, \textit{i.e.,} distinguishing the friends and other individuals and distinguishing different social-tie strengths.  
The agents can further make their own daily schedule autonomously in the city environment, which further leads to social activities, yielding interaction between different agents, including chatting, cooperation, or even conflicts.
Last, the online social network, which is not restricted by the physical space, also provides the environment for social activities. 
The agents can post new content or propagate content of the other users.
Despite the behavior itself, the internal characteristics, including the emotion and attitude of the agent, are also contained in the memory and mechanism of the large language model-based social agents.
Overall speaking, the simulation can be evaluated from both individual-level and population-level perspectives. 
Regarding individual-level simulation, the aim is to generate social behaviors, attitudes, and emotions by leveraging user characteristics and the informational context within social networks.
In the social simulation system S$^3$~\cite{gao2023s} built based on the UGI system, the social agents can accurately simulate the propagation process of information, attitude, and emotion on two representative events about nuclear energy and gender discrimination. 
To summarize, the UGI system provides a good platform to support understanding and simulate social behaviors, including the emerged social phenomenon.

\subsubsection{Decision Making Agent: Task Solving and Personal Assistance}

We present the design cases of decision-making agents in the following two scenarios. 

\textbf{Location recommendation}: 
Location recommendation is one kind of new infrastructure in the area of information overload.
That is, there are too many points of interest (locations) in the city environment, and the individual living there finds it hard to determine where to visit to meet the demands. Furthermore, each individual may have his/her own preferences and interests, which motivates the construction of personalized assistants based on large language model-empowered agents.
In our system, we design an LLM-driven agent for location recommendation based on the LLM's strong ability to understand both user preferences and decision-making. First, the agent can extract critical information from the profile, attributes, and other basic information for a given user. In other words, the LLM agent can be a personal assistant with essential information about the user. Second, the agent is good at planning and scheduling based on the city environment's feedback. Specifically, it is always challenging for a human to directly query or search for locations for visitation since the searching or filtering process will be faced with abundant data and information. To address it, the agent can organize the output of the traditional search and recommendation engines well and even adjust the engine if the results cannot meet the requirements. Last, the agent can communicate well with the user, understand the user's new and instant feedback, and provide textual explanations for recommendation results based on the users' historical behaviors, personal demands, or spatial-temporal context.
In UGI, the large language model-based agents have three major abilities, including 1) understanding the mixed and complex user intents, 2) detecting the user profiles and interests based on historical data and then adjusting recommendations, and 3) identifying the improper user demands given the real-world city environment.

\textbf{Schedule planning}: Effective schedule planning is crucial for efficient daily activity management. Traditional methods, often relying on shortest path algorithms~\cite{eppstein1998finding}, provide time-saving solutions but lack personalization and flexibility. They fail to account for user-specific preferences and cannot dynamically adapt to evolving or abstract requirements, potentially leading to suboptimal scheduling and conflicts. To overcome these limitations, we design a foundation model-driven agent to help users make high quality decisions for nuanced schedule planning. This agent leverages \emph{CityGPT} capabilities, integrating common-sense knowledge to contextualize tasks and offering logical, user-centric planning solutions. Its natural language interface ensures a seamless, intuitive human-computer interaction, significantly enhancing the user experience.

The proposed agent comprises several key components. Upon receiving user’s schedule input, the agent will use the comprehension and reasoning skills of its generative intelligence core to formulate optimal schedule. It will respect the fixed commitments specified by users while accommodating preferences and time constraints. The \emph{memory} module continuously integrates new information, facilitating dynamic adjustments. Through its interfaces for \emph{perceive} and \emph{communicate}, the agent evaluates the feasibility of plans, considering travel time and proximity. It finalizes the schedule with the user preference inferred from \emph{persona} module, ensuring logical coherence and user satisfaction. This approach represents a significant advancement in personalized schedule planning, harnessing foundation model’s deep understanding and reasoning for more tailored and efficient daily organization. 
The design of schedule planning agent represents the basic decision making capability in urban daily life, which can also serve as a useful personal assistance that continuously learns user preference and evolves with urban environment.

\section{Evaluation}
\begin{figure}[t]
    \centering
    \includegraphics[width=1\textwidth]{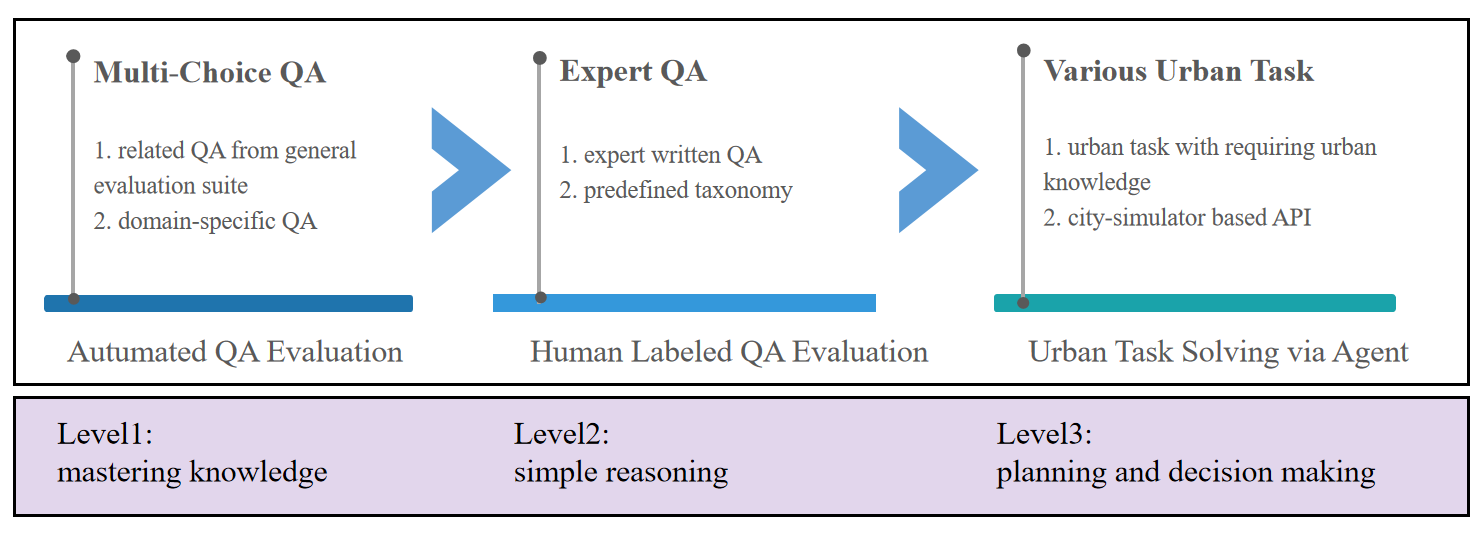}
    \caption{Systematic evaluation framework for urban generative intelligence.}
    \label{fig:eval}
\end{figure}

Here, we introduce a systematic evaluation framework to validate the performance of foundation model and LLM empowered agents for urban generative intelligence. As shown in Figure~\ref{fig:eval}, the whole evaluation framework contains three levels,
\begin{itemize}
  \item \textit{Level 1: evaluating the urban knowledge of CityGPT by automated domain-specific question answering}
  \item \textit{Level 2: evaluating the simple reasoning ability of CityGPT via human labeled domain-specific question answering}
  \item \textit{Level 3: evaluating the planning and decision making ability of generative city agents by solving specific tasks in urban space}
\end{itemize}

\textbf{\textit{Level 1}: Automated domain-specific question answer evaluation.} Evaluation in \textit{Level 1} aims to validate whether the CityGPT really learn and understand the domain-knowledge injected in the training. To do this, we first extract related domain-specific question answer pairs from various general evaluation datasets including Gaokao~\cite{zhang2023evaluating}, CEval~\cite{huang2023c} and CMMLU~\cite{li2023cmmlu} as the first part of evaluation. Besides, we also construct domain-specific question answer pairs based on the pre-training corpus with the help of ChatGPT. Specifically, we first random sample instance from the training corpus. Then, with carefully designed prompts, we require ChatGPT to generate question and answer pairs based on the input context. Finally, we design another prompt to require ChatGPT validate the quality of generated QA and filter the low quality ones. In this way, we can collect questions span diverse topics in the city including transportation, civil engineering, environment, geography and so on. To enable the automated evaluation, all the question answer pairs are formatted as the multiple-choice questions and we use accuracy as the metric.

\begin{figure}[t]
    \centering
    \includegraphics[width=0.42\textwidth]{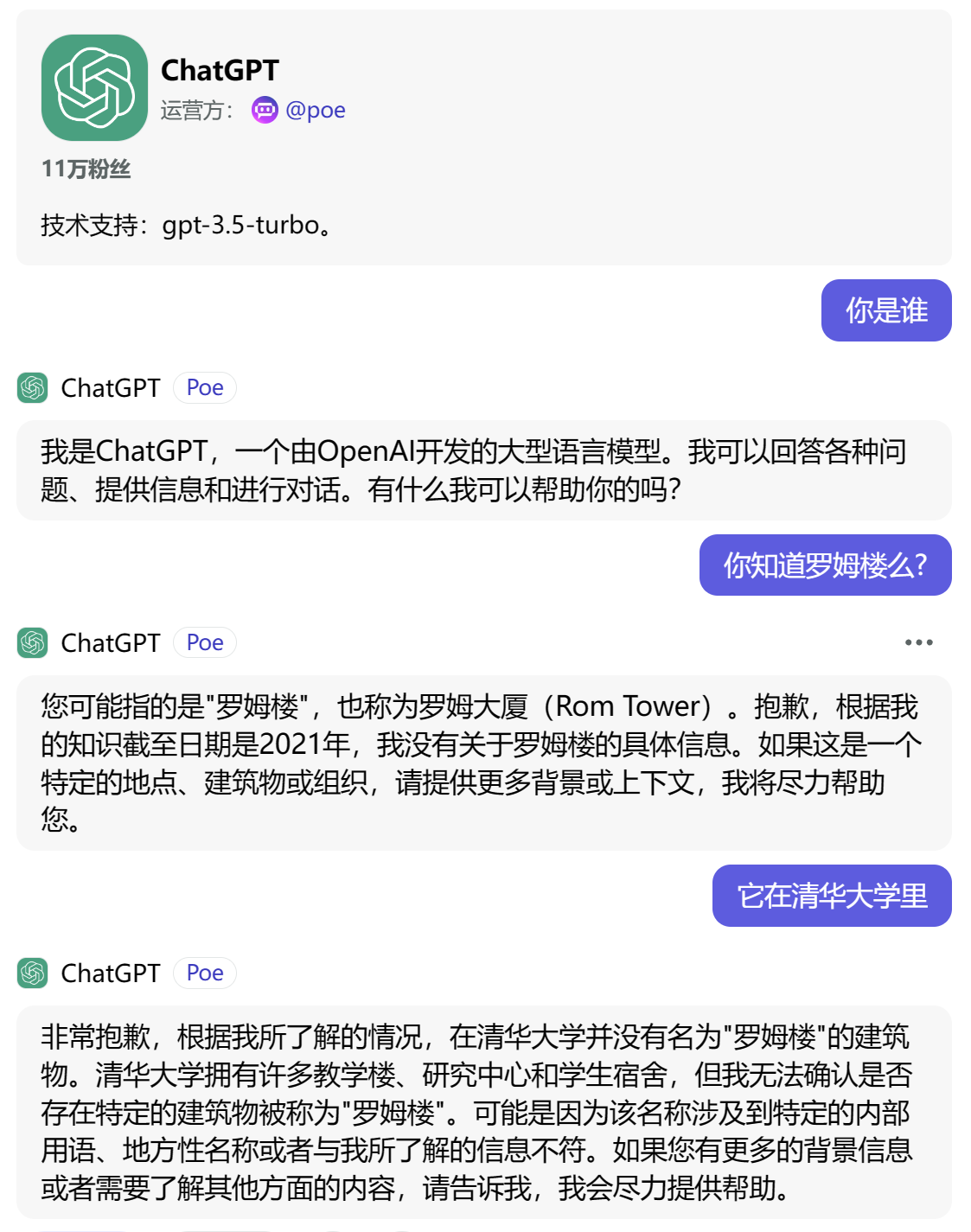}
    \includegraphics[width=0.49\textwidth]{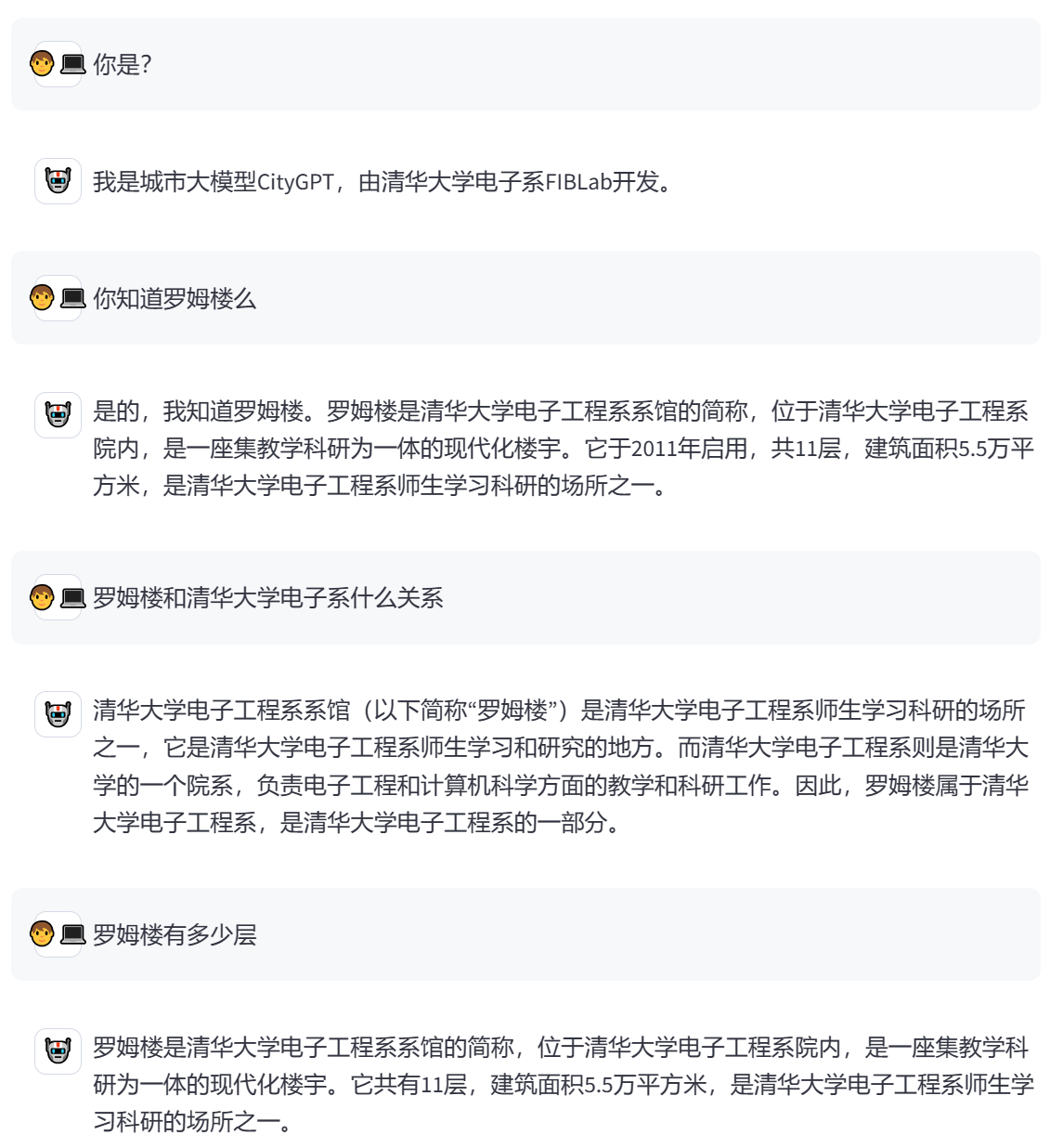}
    \caption{Human labeled domain-specific question answer example on commonsense knowledge. The left is the answer from ChatGPT, the right is the answer from CityGPT.}
    \label{fig:rohm}
\end{figure}

\begin{figure}[t]
    \centering
    \includegraphics[width=0.47\textwidth]{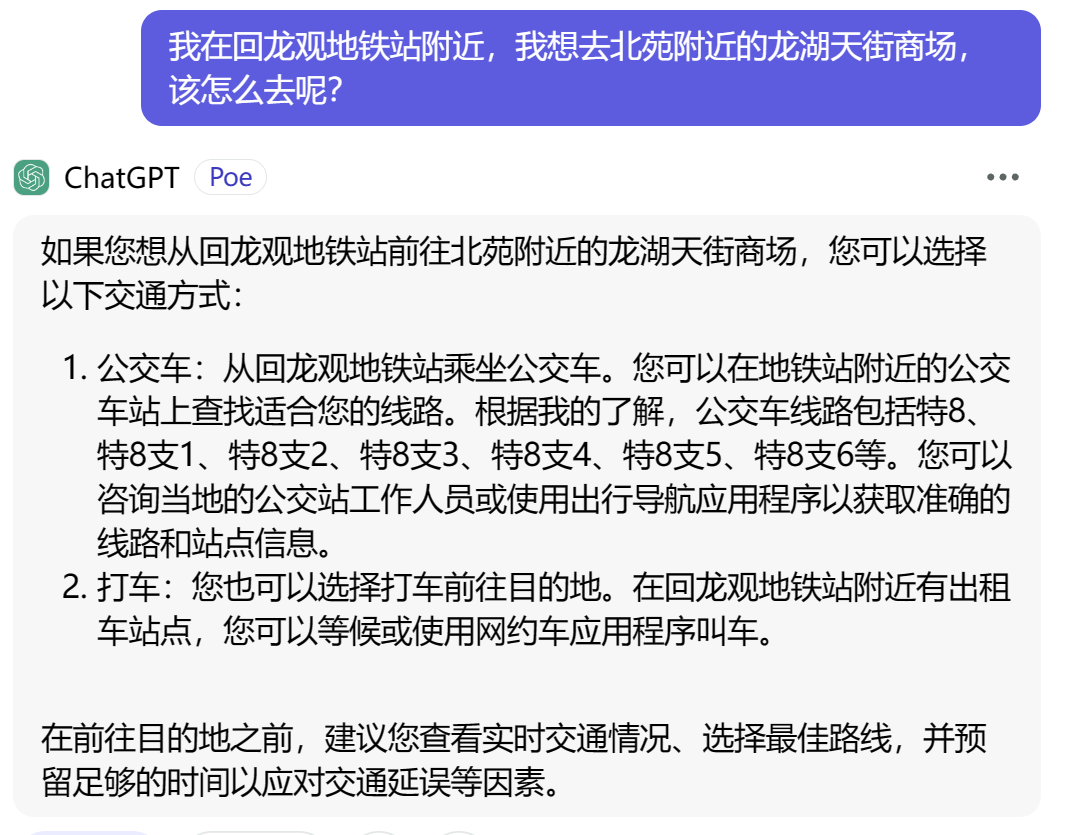}
    \includegraphics[width=0.45\textwidth]{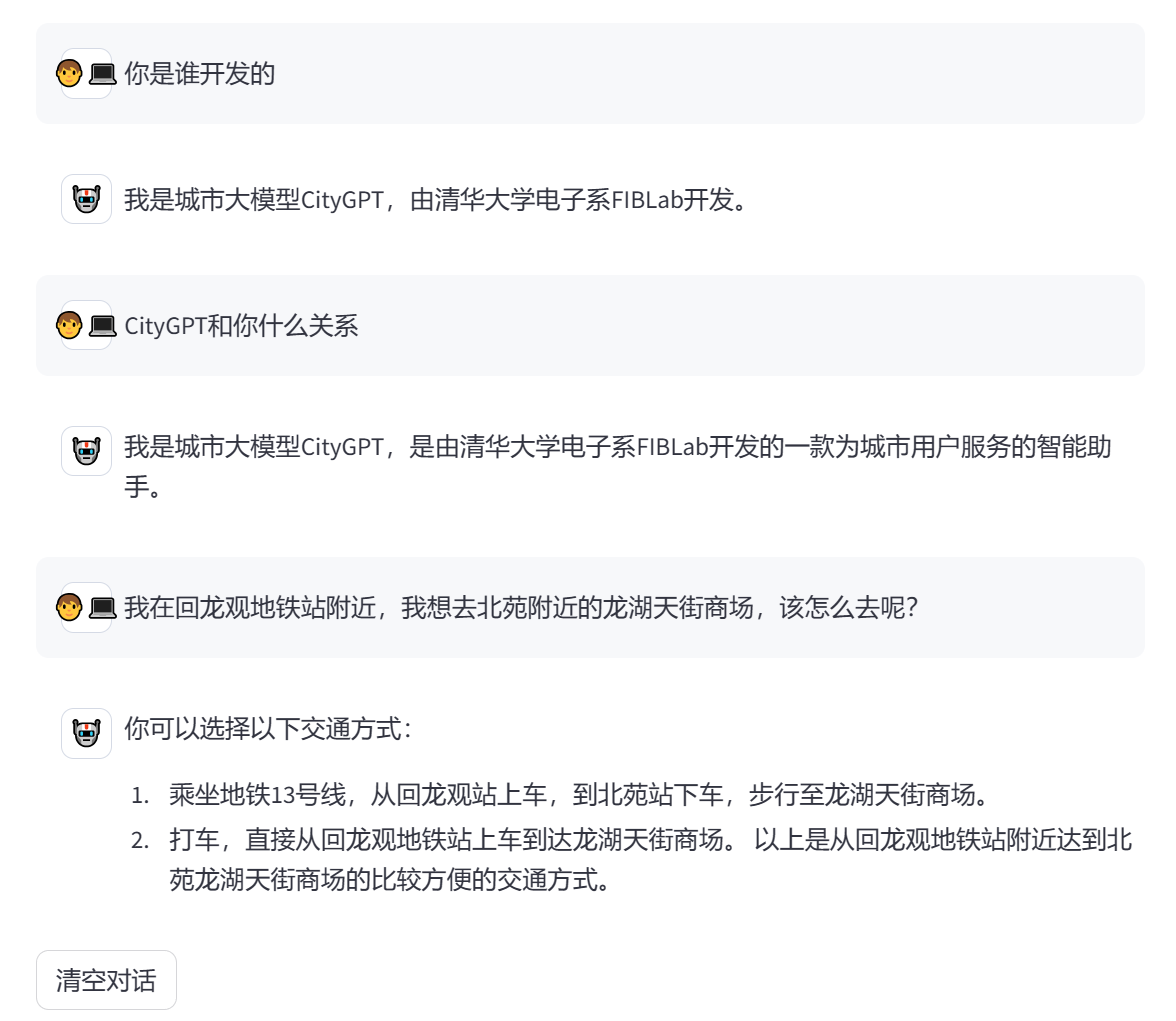}
    \caption{Human labeled domain-specific question answer example on route planning. The left is the answer from ChatGPT, the right is the answer from CityGPT.}
    \label{fig:route}
\end{figure}

\textbf{\textit{Level 2}: Human labeled domain-specific question answer evaluation.} We build \textit{Level 2} evaluation to validate the reasoning ability of CityGPT in the simple scenarios of city. Different from the \textit{Level 1} evaluation, we hope the \textit{Level 2} evaluation can take a step further by answering the questions which can not be directly extracted from the training corpus. Thus, we recruit volunteers with various backgrounds to write new expert problems related to urban intelligence and corresponding answers. To guarantee the coverage of problems on urban space, we predefine a question taxonomy of various domain to lead the topic selection of volunteers. One example of question on common knowledge is shown in Figure~\ref{fig:rohm} and another example on route planning is presented in Figure~\ref{fig:route}. We can find that due to the lack of domain knowledge on urban space, advanced LLMs like ChatGPT cannot solve these simple urban problems. At the same time, CityGPT with domain-knowledge enhancement solve them easily.

\textbf{\textit{Level 3}: Problem solving of various urban tasks.} Finally, we introduce several complicated real-life urban task as \textit{Level 3} evaluation to validate the capability of generative city agents on long term planning and decision making. We use next location prediction, PoI navigation without map, daily schedule planning, and society simulation as four representative tasks. To complete these tasks, beyond the basic ability evaluated before, agents have to master several high-level skills like spatial-temporal reasoning, multi-step goal decomposition, external tool using and so on. During the evaluation, agent is only allowed to access the API and dataset provided by City Simulator. The agents empowered by different foundation models should follow the same structure in the specific task. In each task, we provide hundreds of samples for agents to complete and the overall success rate on these samples are calculated as the final metric of it. It is noted that the definition of success rate various depending on the task. For simulation based tasks, we define the task is successfully completed when its results meet the general law in the field and various pre-defined metrics. 
\section{Enabled Urban Applications}

In this Section, we take several typical examples to discuss how our proposed UGI foundation platform enables to deal with complicated urban tasks and issues from four important urban systems of transportation, business, economy, and society.

\subsection{Transportation System} 

Travel surveys have long been a cornerstone in transportation research, providing indispensable insights into travel behaviors and patterns. These surveys inform urban planning, infrastructure development, and transportation policy, aiding in the creation of more efficient and user-centric transport systems. However, traditional travel surveys, such as household travel surveys and on-board transit surveys, come with significant challenges. They are often expensive and time-consuming to conduct, involving face-to-face interviews, manual data collection, and extensive processing~\cite{stopher2012collecting}. Additionally, the data collected may not adequately capture rapid shifts in travel behavior due to its infrequent nature~\cite{richardson1995survey}. Recent advancements in technology have led to the exploration of data science research in human mobility~\cite{gonzalez2008understanding}. Researchers leverage the increasingly available mobility data to identify the universal rules in human mobility~\cite{song2010modelling}, and design rule-based generator of urban mobility behaviour~\cite{jiang2016timegeo}. This shift is significant as it promises to overcome the limitations of traditional surveys, offering real-time data collection and analysis capabilities.

However, the classic rule-based mobility generator model, such as TimeGeo~\cite{jiang2016timegeo}, leverages simplified statistics rules to simulate individual movements between several frequent locations like home, work, and other. However, they lack a in-depth understanding of mobility intent and user profiles, and hence the travel behaviour they generate are not realistic. Foundation model-driven generative agents bring about important opportunities. The reasoning core of language model possesses not only robust comprehension capabilities for commonsense, but also could make high quality reasoning based on contextual information. In this paper, we describe a generative agent for physical mobility behaviour, which can generate realistic and intention-aware travel behaviour. Such generative model will provide an important opportunity for high-quality and efficient alternative for travel survey.

\subsection{Business Intelligence} 

Business site selection plays a key role in the interdisciplinary areas of urban planning, economic growth, and social development. Traditional site selection methods, often reliant on expert consultants and manual surveys, are resource-intensive and time expensive~\cite{breheny1988practical,kumar2000effect,phelps2018business}. In contrast, research interests have shifted towards a data-driven paradigm, employing machine learning models fed with diverse urban data to evaluate potential sites~\cite{karamshuk2013geo,li2018commercial,liu2019deepstore,xu2016demand}. These models, however, often lack comprehensive feature representation and logical reasoning in their analysis~\cite{yap2018analytic}. Recent advancements have introduced knowledge graphs in business site selection, integrating multifaceted data into a graph structure for enhanced knowledge representation without complex feature engineering~\cite{ji2021survey,liu2021knowledge}. Despite their potential, knowledge graphs face challenges in assimilating varied urban data, refining knowledge for different factors, and ensuring interpretability in decision-making.

Foundation models have emerged as a promising tool, capable of automating text-related tasks with extensive domain knowledge, advanced language generation abilities, and efficient data processing~\cite{wei2022emergent,bubeck2023sparks}. Their application in business site selection offers capabilities like comprehensive information retrieval and real-time decision support. However, foundation models often struggle with accurately recalling facts in knowledge-based content generation~\cite{yang2023chatgpt}. Here, we propose to address these gaps with an integrated intelligent site selection model. It combines the structured knowledge of knowledge graphs with the reasoning and common-sense prowess of language foundation model, which is particularly enhanced for urban problems in \emph{CityGPT}. Utilizing algorithms of reasoning on knowledge graph, this model is designed to deliver precise site selection results with enhanced decision-making quality, clear interpretations and improved efficiency and breadth. Therefore, we can leverage city foundation model to unleash the power in urban knowledge graph and various empirical data to transform business site selection.

\subsection{Urban Economy System}
Agent-based modeling and simulation are of great importance for the research of the economy due to the limitations of other approaches.
Early empirical statistical models, such as the Phelps Model \cite{phelps1967phillips} highlighted in the pioneering works of Hendry~\cite{hendry1982formulation}, delved into data-driven analyses of macroeconomic phenomena. These models unraveled relationships among pivotal variables. Kydland and Prescott \cite{kydland1982time} crafted a computational model geared toward predicting policy outcomes. Later, the advent of Dynamic Stochastic General Equilibrium (DSGE) models \cite{christiano2005nominal} aimed to encapsulate the dynamics of diverse economic variables like output, inflation, consumption, and investment, while accommodating the inherent uncertainty and randomness within economic processes. However, as pointed out by Farmer~\cite{farmer2009economy}, these models operate under the assumption of a perfect world, which motivates agent-based modeling and simulation for the economy.
That is, the large language model-based economic simulation based on our system can be an environment to deploy various relevant applications.

\noindent \textbf{Macroeconomic behavior.}
This simulation offers an ideal platform to scrutinize and emulate complex macroeconomic behaviors. By leveraging the interplay of diverse agents and institutions, the model can elucidate emergent behaviors, market dynamics, and the ripple effects of economic decisions. Understanding these behaviors is crucial for forecasting economic trends and devising resilient strategies in response to varying scenarios.

\noindent \textbf{Macroeconomic activities.}
Through this simulation framework, the intricate landscape of macroeconomic activities can be explored comprehensively. From trade dynamics and investment patterns to consumption trends and labor market behaviors, the model provides a simulated environment to examine and evaluate diverse economic activities.

\noindent \textbf{Policy making.}
This simulation serves as an invaluable tool for policymakers to test and assess the efficacy of different policy interventions in a controlled environment. By simulating policy scenarios and their potential impacts on various economic indicators, policymakers can fine-tune strategies, evaluate trade-offs, and anticipate unintended consequences before implementation. This proactive approach to policymaking helps in devising robust, adaptive policies conducive to sustainable economic growth.

In essence, our system's large language model-based economic simulation platform offers a versatile and robust framework for investigating, understanding, and shaping macroeconomic behavior, activities, and policy outcomes.

\subsection{Urban Society}

Understanding our society is the core of social sciences, for which the proposal and validation of theory highly relies on social experiments. Due to the high cost of real-world social experiments, the simulation is a very promising approach.
There are two key perspectives in social simulation, as outlined by Gilbert~\cite{gilbert2005simulation}: 1) the dynamic interaction among individuals, and 2) the status evolving of the population.  By simulating social activities, both researchers and practitioners gain the ability to forecast the future progression of individual behaviors and the overall status of populations. Moreover, these simulations provide experimental arenas where interventions can be implemented and their effects observed.
The applications supported by our system and agents can be summarized as follows.

\noindent \textbf{Understanding individual behaviors in society.}
The simulation system, along with the LLM-driven agents, enables a deep dive into individual behaviors within social contexts. By emulating these behaviors, researchers and practitioners can forecast and comprehend how individual actions are driven by internal mechanisms and external contexts or factors.

\noindent \textbf{Predicting population-level dynamics.}
Beyond individual behaviors, the system facilitates the prediction of broader population dynamics. It offers insights into how collective behaviors, trends, and group interactions evolve over time, aiding in anticipating societal shifts and trends.

\noindent \textbf{Experimenting with interventions and policy evaluation.}
The simulation serves as an experimental ground for testing interventions in simulated social environments. Researchers and practitioners can implement and study the impact of various interventions, policies, or changes within these controlled settings, providing crucial insights into potential real-world outcomes. Thus, policymakers can develop and evaluate policies in a simulated societal landscape. By testing proposed policies virtually, they can assess their potential effects and fine-tune strategies before real-world deployment.

\noindent \textbf{Emergency and risk management.}
In the real-world scenario, emergency-related data is always sparse, which leads to the challenge of risk prevention. 
 By exploring different potential outcomes based on varying parameters, the government can prepare strategies to mitigate risks and adapt to changing circumstances, supported by the simulation system and LLM-driven agents in simulated society.

\section{Discussion}
We discuss the open challenges and important future research directions of urban generative intelligence platform from the following aspects.

\subsection{Dive into Complicated Urban Issues} 

Urban environments are dynamic and multifaceted, which are increasingly confronted with a myriad of complex issues stemming from their intricate networks encompassing physical, social, economic, and environmental factors. As mentioned in the Introduction, the rapid urbanization exacerbates challenges like traffic congestion, environmental pollution, resource scarcity, and infrastructure strain, alongside socio-economic issues like social inequality and housing crises. Addressing these issues is crucial for sustainable, equitable urban development and maintaining the vitality of cities in a global context.

To navigate these complexities, our proposed foudation platform of Urban Generative Intelligence (UGI) can foster emergent and sophisticated urban solutions. By leveraging the multi-source urban data and creating a real urban environment for interaction beyond traditional sandboxes or virtual simulations, with the LLM-empowered embodied agents, UGI enables deep, context-aware insights, offering nuanced understandings of urban dynamics. Morevoer, UGI allows for the emergence of intelligent solutions, which is able to address complex urban issues through advanced cognitive capabilities similar to human intelligence, while with the power of computational intelligence.

While UGI holds promise in tackling urban complexities, several challenges necessitate further exploration. Bridging the gap between advanced technological capabilities and practical, real-world urban applications remains a crucial hurdle. This includes adapting UGI to rapidly evolving urban dynamics and policy landscapes. Moreover, there is a pressing need to develop advanced embodied agent for a more nuanced, systematic understanding of urban complexities, integrating the diverse social, economic, and environmental aspects of urban life. Additionally, adapting these solutions to the escalating challenges of rapid urbanization and climate change is vital for ensuring sustainable and resilient urban development. Addressing these problems is critical for the successful implementation and evolution of UGI, making it a truly transformative tool for urban problem-solving.

\subsection{Scale Up to Large City}

Recent advancements in LLMs have opened new frontiers in simulating complex urban systems. Studies reveal that LLM agents, when personalized with diverse roles such as executives, engineers, and designers, can synergistically solve complex tasks like software development, making significant strides in designing, coding, testing, and documentation processes~\cite{qian2023communicative}. The scalability of these simulations, introducing more varied personas, has been shown to be beneficial across various domains~\cite{zhuge2023mindstorms}, which are particularly of simulating large urban systems. 

However, simulating societies of large-scale LLM agent, reflecting the complex constraints in urban environments, faces substantial computational challenges. Research efforts are geared towards optimizing the memory footprint and operational efficiencies of these models~\cite{sheng2023high,aminabadi2022deepspeed}. Techniques like model compression through knowledge distillation and quantization have been proposed~\cite{zhu2023survey}. Specifically, in urban simulations, batch prompting has emerged as a crucial technique, enhancing efficiency by simulating multiple agents concurrently, showing up to a 5x improvement in inference time and cost~\cite{cheng2023batch}. Moreover, the MetaGPT framework, initially applied in virtual software companies, presents a promising approach for efficient multi-agent collaboration in urban simulations~\cite{hong2023metagpt}. Its shared message pool and subscription mechanism offer significant reductions in resource consumption. Despite these advancements, simulating expansive urban societies with LLM agents remains a formidable challenge, limiting the full potential of these simulations. Successfully simulating large-scale urban environments with LLM agents could not only enhance performance in specific tasks but also mimic emergent properties of human societies, offering insights into complex urban dynamics~\cite{caldarelli2023role}. Thus, achieving full-process acceleration in LLM agent simulations remains a critical, yet unresolved, task in urban science.

\subsection{Openness of the Environment}

As a foundational platform that integrates advanced technologies such as big data, simulation, and LLMs, UGI's capabilities are not limited to providing realistic urban environments.
With UGI's open capabilities, users can transform their environments at will based on real cities, or even create a new city.
Specifically, users can adjust AOI, POI and other data to change the urban spatial structure, land use type, so as to change the spatial distribution of urban functions.
Based on the new urban spatial structure and functional distribution, users can use existing algorithms~\cite{rong2023interdisciplinary,rong2023complexity,yuan2022activity,yuan2023learning} to generate urban human activities under new conditions.
Users are also allowed to create human activities using their own algorithms or data sources.
Besides, the city's road network, infrastructure networks, and even the image data, are all open and allow users to make any modifications on the copy that belongs to them.
Through the openness of the environment, we hope that UGI will not only be used to build LLM-based agents in the city, but also that it will be able to provide a full range of intelligence for the planning, design, and governance of future cities, and promote multidisciplinary paradigm innovation in the urban field.

\subsection{Developer Community}

As a topic that integrates the latest achievements in big data, urban simulation, LLMs and other fields, the development of UGI requires the collaboration of researchers and developers in multiple fields.
This requires the establishment of a multi-disciplinary collaborative UGI developer community.
In the community, researchers in the field of big data can share their data sets, data processing methods, and data generation methods to provide high-quality data streams for UGI.
People interested in urban simulation can add new functions to the open infrastructure, improve its computing performance, and design more reasonable interfaces.
Large language model researchers can provide insights for the training of \textit{CityGPT}.
Researchers in urban-related fields, such as urban planning, traffic management, economics, etc., can build their own agents that solve domain-specific problems through programming or natural language interface.
The community will be a highly interdisciplinary community that will inspire many interesting ideas and research questions, help solve urban problems, and achieve smart and sustainable urban development.

\section{Conclusion}\label{sec::conclusion}
In conclusion, Urban Generative Intelligence (UGI) marks a significant advancement in the field of city science and urban computing, bridging the gap between cutting-edge technological capabilities and practical urban system applications. By innovatively integrating Large Language Models (LLMs) with urban data and digital twins, UGI provides a nuanced, dynamic platform for the development and deployment of embodied agents with human-level intelligence. These agents, empowered by CityGPT, are adept at addressing diverse urban challenges, offering insights and solutions across social, economic, and environmental dimensions. This foundational platform not only propels forward the field of urban science but also sets new paradigm of generative intelligence in urban space. UGI's comprehensive approach to modeling complex urban systems heralds a new era of intelligent, sustainable, and resilient urban development, paving the way for future cities that are more adaptive and responsive to the evolving needs of their inhabitants.

\bibliographystyle{plain}
\bibliography{bibliography}

\end{document}